\documentclass{article}
\usepackage[square,numbers]{natbib}

\usepackage[eandd, final]{neurips_2026}

\usepackage[utf8]{inputenc} 
\usepackage[T1]{fontenc}    
\usepackage{hyperref}       
\usepackage{url}            
\usepackage{booktabs}       
\usepackage{amsfonts}       
\usepackage{nicefrac}       
\usepackage{microtype}      
\usepackage{xcolor}         
\usepackage{graphicx} 
\usepackage{listings}
\usepackage[most]{tcolorbox}
\usepackage{booktabs}        
\usepackage{multirow}        
\usepackage{array}           
\usepackage{tabularx}
\usepackage{colortbl}        
\usepackage{siunitx}         
\usepackage{caption}
\usepackage{tikz}            
\usetikzlibrary{shapes.geometric, positioning}

\usepackage{enumitem}
\usepackage{wrapfig}
\usepackage[table,dvipsnames]{xcolor}
\usepackage{hyperref}
\usepackage{booktabs}
\usepackage{tabularx}
\usepackage{subcaption}
\usepackage{placeins}
\usepackage{array}
\usepackage{url}
\usepackage{multirow}
\hypersetup{
  colorlinks=true,
  urlcolor=Sepia,       
  linkcolor=black,
  citecolor=black
}
\definecolor{tPres}{HTML}{555B8E}   
\definecolor{tWeb}{HTML}{556F95}    
\definecolor{tPos}{HTML}{5494A1}    
\definecolor{tData}{HTML}{7EB2AC}   
\definecolor{tRep}{HTML}{B1CEB5}    
\definecolor{linkgray}{HTML}{6B6B6B} 
\colorlet{currentSkillColor}{gray}
\newcommand{\sk}[2]{\href{#1}{\textcolor{currentSkillColor}{\scriptsize\texttt{#2}}}}
\definecolor{DivPosText}{HTML}{2E6F4E}   
\definecolor{DivNegText}{HTML}{9C3848}   
\definecolor{DivPosBg}{HTML}{EAF0F7}     
\definecolor{DivNegBg}{HTML}{FCF4EF}     
\newcommand{\cvalpos}[2]{\cellcolor{DivPosBg}\cval{#1}{#2}}
\newcommand{\cvalneg}[2]{\cellcolor{DivNegBg}\cval{#1}{#2}}

\newcommand{\divpos}[1]{\cellcolor{DivPosBg}\textcolor{DivPosText}{#1}}
\newcommand{\divneg}[1]{\cellcolor{DivNegBg}\textcolor{DivNegText}{#1}}

\definecolor{HeaderBlue}{HTML}{1F3A5F}    
\definecolor{HeaderText}{HTML}{FFFFFF}
\definecolor{StripeRow}{HTML}{F5F2EC}     
\definecolor{TaskBand}{HTML}{E5E3DC}      
\definecolor{Best}{HTML}{1D6F47}          
\definecolor{Worst}{HTML}{8C2D2D}         
\definecolor{Rule}{HTML}{B4B2A9}
\definecolor{TextMuted}{HTML}{6A6A65}

\newcommand{\iconBase}[1]{%
  \raisebox{0.3ex}{%
    $\vcenter{\hbox{%
      \begin{tikzpicture}[x=1.5ex, y=1.5ex, line width=0.5pt]
        \useasboundingbox (0,0) rectangle (2.0,2.0);
        #1
      \end{tikzpicture}%
    }}$%
  }%
}
\newcommand{\iconReport}{\iconBase{%
  \draw[HeaderBlue!85] (0.10,0.30) rectangle (1.90,1.70);
  \fill[HeaderBlue!85] (0.30,1.30) rectangle (1.50,1.45);
  \fill[HeaderBlue!85] (0.30,1.00) rectangle (1.20,1.15);
  \fill[HeaderBlue!85] (0.30,0.70) rectangle (1.50,0.85);
  \fill[HeaderBlue!85] (0.30,0.40) rectangle (1.30,0.55);
}}
\newcommand{\iconDataViz}{\iconBase{%
  \fill[HeaderBlue!85] (0.30,0.50) rectangle (0.60,1.15);
  \fill[HeaderBlue!85] (0.70,0.50) rectangle (1.00,1.55);
  \fill[HeaderBlue!85] (1.10,0.50) rectangle (1.40,0.95);
  \fill[HeaderBlue!85] (1.50,0.50) rectangle (1.80,1.70);
  \draw[HeaderBlue!85] (0.20,0.50) -- (1.85,0.50);
}}

\newcommand{\iconPPT}{\iconBase{%
  \draw[HeaderBlue!85, line width=0.5pt] (0.10,0.45) rectangle (1.90,1.55);
  \fill[HeaderBlue!85] (0.10,1.32) rectangle (1.90,1.55);
  \fill[HeaderBlue!40] (0.20,0.60) rectangle (0.90,1.20);
  \fill[HeaderBlue!85] (1.05,1.10) rectangle (1.78,1.20);
  \fill[HeaderBlue!85] (1.05,0.92) rectangle (1.65,1.02);
  \fill[HeaderBlue!85] (1.05,0.74) rectangle (1.78,0.84);
  \fill[HeaderBlue!85] (1.05,0.60) rectangle (1.50,0.66);
}}

\newcommand{\iconPoster}{\iconBase{%
  \draw[HeaderBlue!85] (0.20,0.20) rectangle (1.80,1.80);
  \fill[HeaderBlue!40] (0.30,1.05) rectangle (1.70,1.70);
  \fill[HeaderBlue!85] (0.30,0.75) rectangle (1.70,0.90);
  \fill[HeaderBlue!85] (0.30,0.50) rectangle (1.40,0.65);
  \fill[HeaderBlue!85] (0.30,0.25) rectangle (1.50,0.40);
}}

\newcommand{\iconWeb}{\iconBase{%
  \draw[HeaderBlue!85] (0.20,0.45) rectangle (1.80,1.55);
  \fill[HeaderBlue!85] (0.20,1.30) rectangle (1.80,1.55);
  \fill[red!75]         (0.35,1.42) circle (0.10);
  \fill[orange!90]      (0.55,1.42) circle (0.10);
  \fill[green!60!black] (0.75,1.42) circle (0.10);
  \fill[HeaderBlue!85] (0.35,0.95) rectangle (1.20,1.08);
  \fill[HeaderBlue!85] (0.35,0.72) rectangle (1.45,0.85);
  \fill[HeaderBlue!85] (0.35,0.49) rectangle (1.05,0.62);
}}

\newcommand{\cval}[2]{$#1\pm#2$}
\newcommand{\bcval}[2]{$\mathbf{#1\pm#2}$}

\newcommand{\taskband}[3]{%
  \rowcolor{TaskBand}%
  \multicolumn{11}{l}{%
    \hspace*{-4pt}%
    \rule{0pt}{2.6ex}%
    #1\,\,\textbf{#2}\hspace{0.6em}\textcolor{TextMuted}{\small\textit{#3}}%
  }\\
}

\newcommand{\taskbandskill}[3]{%
  \rowcolor{TaskBand}%
  \multicolumn{9}{l}{%
    \hspace*{-4pt}%
    \rule{0pt}{2.6ex}%
    #1\,\,\textbf{#2}\hspace{0.6em}\textcolor{TextMuted}{\small\textit{#3}}%
  }\\
}

\definecolor{schemaBorder}{HTML}{1F3A93}
\definecolor{schemaTitle}{HTML}{1F3A93}
\definecolor{schemaBg}{HTML}{F7F9FC}
\definecolor{jsonKey}{HTML}{1F4FA0}
\definecolor{jsonStr}{HTML}{0B7A45}
\definecolor{jsonNum}{HTML}{A0410B}
\definecolor{jsonComment}{HTML}{7A7A7A}
\definecolor{jsonPunct}{HTML}{2B2B2B}

\lstdefinelanguage{jsonc}{
  sensitive=true,
  morestring=[b]",
  morecomment=[l]{//},
  morecomment=[s]{/*}{*/},
  literate=
    *{:}{{{\color{jsonPunct}{:}}}}{1}
     {,}{{{\color{jsonPunct}{,}}}}{1}
     {\{}{{{\color{jsonPunct}{\{}}}}{1}
     {\}}{{{\color{jsonPunct}{\}}}}}{1}
     {[}{{{\color{jsonPunct}{[}}}}{1}
     {]}{{{\color{jsonPunct}{]}}}}{1}
     {true}{{{\color{jsonNum}true}}}{4}
     {false}{{{\color{jsonNum}false}}}{5}
     {null}{{{\color{jsonNum}null}}}{4},
}

\lstdefinestyle{schemaJson}{
  language=jsonc,
  basicstyle=\ttfamily\footnotesize,
  keywordstyle=\color{jsonKey}\bfseries,
  stringstyle=\color{jsonStr},
  commentstyle=\color{jsonComment}\itshape,
  showstringspaces=false,
  breaklines=true,
  breakatwhitespace=true,
  columns=fullflexible,
  keepspaces=true,
  tabsize=2,
  upquote=true,
  numbers=none,
  frame=none,
  aboveskip=2pt, belowskip=2pt,
}

\newtcolorbox{schemabox}[2][]{%
  enhanced, breakable,
  colback=schemaBg,
  colframe=schemaBorder,
  boxrule=0.7pt,
  arc=2.5pt, outer arc=2.5pt,
  fonttitle=\bfseries\sffamily,
  coltitle=white,
  title={#2},
  attach boxed title to top left={xshift=10pt, yshift*=-2pt},
  boxed title style={
    colback=schemaTitle,
    colframe=schemaTitle,
    arc=2pt, outer arc=2pt,
    boxrule=0pt,
    top=2pt, bottom=2pt, left=6pt, right=6pt,
  },
  top=14pt, bottom=6pt, left=6pt, right=6pt,
  before skip=12pt, after skip=12pt,
  #1
}

\definecolor{maincolor}{RGB}{75,85,99}      
\definecolor{bgcolor}{RGB}{249,250,251}     
\definecolor{titlebg}{RGB}{107,114,128}     

\newtcolorbox{scorebox}[1]{
  enhanced, breakable,
  colback=bgcolor, colframe=maincolor!40,
  boxrule=0.5pt, arc=3pt,
  left=10pt, right=10pt, top=10pt, bottom=6pt,
  title=#1,
  fonttitle=\bfseries\small,
  coltitle=white,
  colbacktitle=titlebg,
  attach boxed title to top left={xshift=10pt, yshift=-8pt},
  boxed title style={colframe=titlebg, arc=2pt, boxrule=0pt,
                     top=2pt, bottom=2pt, left=6pt, right=6pt},
  before skip=12pt, after skip=4pt,
}

\definecolor{rubricBorder}{HTML}{0F6E5C}
\definecolor{rubricTitle}{HTML}{0F6E5C}
\definecolor{rubricBg}{HTML}{F4FBF8}
\definecolor{rubricTag}{HTML}{0F6E5C}

\lstdefinestyle{promptText}{
  basicstyle=\ttfamily\footnotesize,
  showstringspaces=false,
  breaklines=true,
  breakatwhitespace=true,
  columns=fullflexible,
  keepspaces=true,
  upquote=true,
  numbers=none,
  frame=none,
  aboveskip=2pt, belowskip=4pt,
  xleftmargin=2pt, xrightmargin=2pt,
}

\newtcolorbox{rubricbox}[2][]{%
  enhanced, breakable,
  colback=rubricBg,
  colframe=rubricBorder,
  boxrule=0.7pt,
  arc=2.5pt, outer arc=2.5pt,
  fonttitle=\bfseries\sffamily,
  coltitle=white,
  title={#2},
  attach boxed title to top left={xshift=10pt, yshift*=-2pt},
  boxed title style={
    colback=rubricTitle, colframe=rubricTitle,
    arc=2pt, outer arc=2pt, boxrule=0pt,
    top=2pt, bottom=2pt, left=6pt, right=6pt,
  },
  top=14pt, bottom=6pt, left=6pt, right=6pt,
  before skip=12pt, after skip=12pt,
  #1
}

\newcommand{\rubric}[2]{%
  \par\medskip\noindent
  \colorbox{rubricTag}{\textcolor{white}{\sffamily\bfseries\small\,#1\,}}\quad
  \textsf{\textbf{\small #2}}\par\smallskip
}

\title{OpenSkillEval: Automatically Auditing the Open Skill Ecosystem for LLM Agents}

\author{
Jiahao Ying\textsuperscript{1}\thanks{Equal contribution.},
Boxian Ai\textsuperscript{2}\footnotemark[1],
Wei Tang\textsuperscript{3},
Siyuan Liu\textsuperscript{2},
Yixin Cao\textsuperscript{2}\thanks{Corresponding Author}\\
\textsuperscript{1}Singapore Management University \\
\textsuperscript{2}Institute of Trustworthy Embodied AI, Fudan University \\
\textsuperscript{3}Joy Future Academy, JD \\
}
\begin{document}

\maketitle

\begin{abstract}
Skills, i.e., structured workflow instructions distilled for large language models (LLMs), are becoming an increasingly important mechanism for improving agent performance on real-world downstream tasks. However, as the open-source skill ecosystem rapidly expands, it remains unclear how different models and agent frameworks interact with skills, how to evaluate skill quality, and how users should select skills under practical cost-performance trade-offs. In this paper, we present \textsc{OpenSkillEval}, an automatic evaluation framework for both skill-augmented agent systems and the skills themselves. Instead of relying on static benchmarks, \textsc{OpenSkillEval} automatically constructs realistic task instances from evolving real-world artifacts across five categories of downstream applications: presentation generation, front-end web design, poster generation, data visualization, and report generation. It further collects and organizes community-contributed skills for controlled comparison under unified task settings. Using more than 600 dynamically generated task instances and 30 open-source skills, we conduct a systematic evaluation of state-of-the-art models and agent frameworks. Our results show that skill availability does not guarantee effective skill usage, that the benefit of skill augmentation depends strongly on both the underlying model and the agent framework, and that many publicly popular skills do not consistently outperform base agents without skills. These findings highlight the need for dynamic, task-grounded evaluation and provide practical insights into the design, selection, and deployment of skills for LLM agents. Additional cases and benchmark resources are available on the project website: \url{https://yingjiahao14.github.io/OpenSkillEval-Web/}.
\end{abstract}

\section{Introduction}

Recent advances in increasingly capable large language models (LLMs)~\cite{openai2026gpt54, anthropic2026opus46}, together with the rapid development of agent client frameworks~\cite{claudecode2025,openaicodex2025}, have created a promising opportunity to deploy models as autonomous agents for complex downstream tasks, including report generation, document management, and web design. However, because agentic tasks are often open-ended, the overall behavior of an agent can be difficult to predict, and in many challenging settings the agent's intrinsic capability alone may be insufficient for reliable task completion. To better enable models to handle such structured yet complex workflows, developers often formalize personal experience or accumulated best practices into explicit procedures and distill them into structured instructions for agents. These workflow-oriented, formatted instructions are commonly referred to as skills~\cite{anthropic2025skills}. 

Given the promise of skills for augmenting agent capabilities in downstream task completion, strong community participation over the past few months has led to the creation and integration of a large number of skills into the ecosystem for LLM agents.
However, this rapid growth has also introduced several important challenges.
First, it remains unclear how different agent frameworks perform on downstream tasks in general, and how they interact with the added skill during execution. Lack of systematic evaluation makes it difficult to assess their actual effectiveness. As a result, users may struggle to choose suitable agents for downstream tasks. If an agent lacks the capability to properly execute a provided skill, the practical value of skill augmentation can be greatly reduced. On the other hand, if an agent is already sufficiently strong to solve the task on its own, introducing skills may bring only limited benefit while still increasing execution cost.
Second, individually distilled skills may reflect only partial or subjective experience, which can limit their generalizability. As the number of available skills continues to grow, it also becomes increasingly unclear how to select the most appropriate skills for prompting, especially when users must balance performance and cost. In addition, the repeated submission of redundant or low-quality skills imposes substantial maintenance overhead on the community and contributes to the bloating of the overall ecosystem.

To address these challenges, we propose \textbf{OpenSkillEval}, an automatic evaluation framework for both skill-augmented agent systems and the skills themselves in downstream applications. Instead of relying on static benchmarks, OpenSkillEval dynamically generates test cases that continuously reflect evolving user needs, enabling a more realistic, timely, and comprehensive evaluation setting. Based on these dynamically constructed task instances, we evaluate the effectiveness and efficiency of different models and agent frameworks, both with and without skill augmentation, across diverse downstream tasks. Moreover, by collecting skill sets from the open-source community for each target application, OpenSkillEval enables controlled comparisons of different skills under the same task setting, making it possible to analyze their relative quality, robustness, and transferability. 
Across five major real-world application categories --- presentation generation, front-end web design, poster generation, data visualization, and report generation --- and more than 600 tasks, we evaluate a range of state-of-the-art agent architectures and derive several key findings:
\textbf{1)} We find that the presence of a skill does not guarantee that an agent will actually use it effectively. Across different client frameworks, agents explicitly read the provided skill in only about 48\% of cases on average under a realistic skill-access setting. and in many cases do not faithfully follow the provided skill instructions at all. 
This suggests that in realistic online settings, where the context is more complex and noisy, carefully designed skills may still be under-utilized or even ignored by the agent (Section~\ref{subsec: trajectory analysis}); \textbf{2)} We observe substantial differences across agent frameworks in how effectively they leverage skill augmentation. In some cases, a weaker base model can achieve performance comparable to that of a stronger model when paired with well-designed skills and a suitable framework. However, when the underlying model is intrinsically weak at solving a task, simply adding skills does not reliably produce meaningful gains (Section~\ref{subsec: model comparison}); 
\textbf{3)} Skill quality varies substantially across open-source skills: richer and better-designed priors can help agents trade increased input scaling for improved performance, but many popular skills still fail to outperform base agents while introducing additional cost. Our analysis further provides practical takeaways for skill format and design (Section~\ref{subsec: skills analysis}).

\section{OpenSkillEval Framework} \label{sec: framework}

\begin{figure*}[htp]
\centering
     \includegraphics[scale=0.42]{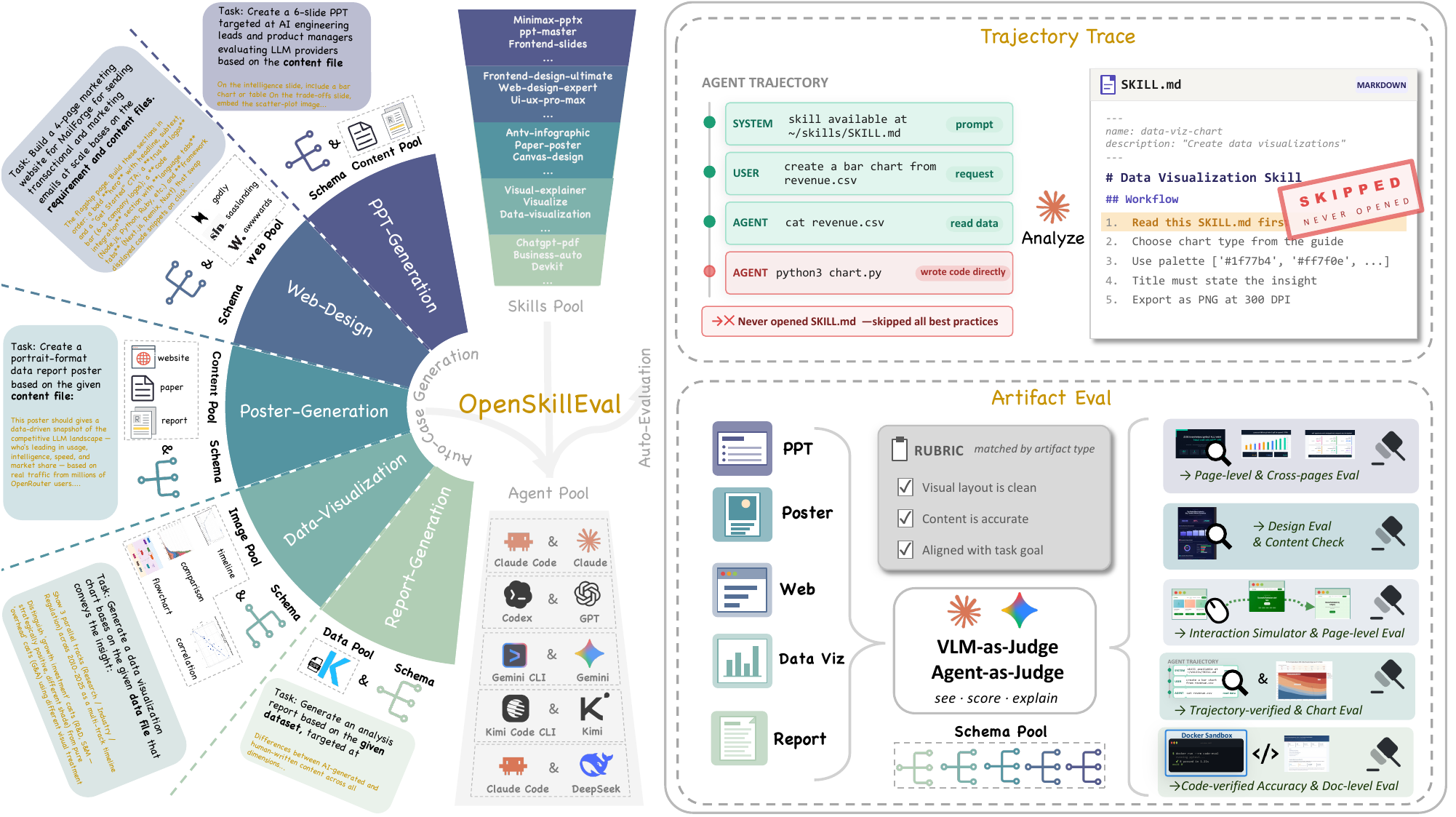}
     \caption{Overview of the {OpenSkillEval} framework. The framework supports automatic test case generation for five core task categories by reflecting evolving user needs. It further enables automatic evaluation from two complementary perspectives: (1) analysis of model trajectory traces to study how skills are used within skill-augmented agent systems, and (2) assessment of the quality of the final artifacts produced under skill augmentation.}

     \label{fig:framework}

\end{figure*}

To effectively evaluate the rapidly growing and increasingly bloated open skill ecosystem for LLM agents, we design {OpenSkillEval} as a sustainable and maintainable framework for real-world downstream application tasks. Accordingly, we decompose the system into several core components. The first component is an automatic test case generation pipeline, which constructs representative evaluation instances for target downstream tasks grounded in realistic user needs (Section~\ref{subsec: auto-case generation}). The second component is a skill collection and organization pipeline, which curates skills from the open-source community and organizes them by task category to support continuous tracking of skill development (Section~\ref{subsec: skills collection}). The final component is an evaluation pipeline that systematically measures the effectiveness of skills in downstream settings. This evaluation considers not only the quality of the final generated artifacts, but also the intermediate agent trajectories, which provide insights into whether and how an agent appropriately invokes and applies skills during task execution in real-world task environments (Section~\ref{subsec: auto-evaluation}). Based on these components, {OpenSkillEval} enables efficient and dynamic auditing of the open skill ecosystem from two complementary perspectives: agent-level comparison and skill-level comparison. Figure~\ref{fig:framework} presents an overview of the framework. In the following subsections, we describe the implementation of each component.

\subsection{Automatic Case Generation} \label{subsec: auto-case generation}
To ensure the practical relevance and feasibility of our evaluation, we focus on five categories of commonly used real-world downstream tasks: presentation generation, front-end web design, poster generation, data visualization, and report generation. We select these tasks because they are representative of common real-world applications, require nontrivial multi-step reasoning and tool use, and typically produce concrete artifacts that can be directly evaluated.
To construct test cases that reflect realistic user needs in these scenarios, we adopt an \textbf{Artifact-driven Case Generation Strategy}. Instead of manually writing task instructions from scratch, we begin with existing high-quality artifacts and infer the underlying user requests that could have led to their creation. This reverse construction process allows us to build evaluation instances that are more closely aligned with real-world usage patterns and expected outputs.
Concretely, each task category instantiates this artifact-driven strategy through a three-stage pipeline. First, we automatically collect artifacts or source materials $(S)$ from diverse external repositories as the content basis for reconstructing realistic user intents. Second, we perform task extraction, where LLMs analyze the collected materials under predefined schemas to construct structured task specification $(T)$ together with its corresponding natural-language instruction $(I)$. Here, the task specification captures the underlying user intent, constraints, and expected outputs implied by the artifact or its supporting context. Third, we apply a validation procedure to verify that the extracted task specification is internally consistent and compatible with the source content, ensuring that the resulting task instances are both coherent and realistic. Below, we describe the source selection and extraction process for each task category.

\textbf{Presentation Generation.}
For presentation (hereafter \textsc{PPT}) generation, we collect publicly available, up-to-date, and information-dense webpages and documents, such as benchmark leaderboards, industry reports, open-data portals, and academic papers, that naturally lend themselves to slide-deck summarization and restructuring. Each source is crawled, snapshot-preserved, and then processed through an LLM-based pipeline to produce a slide-level task specification with explicit per-slide content goals, thereby forming realistic presentation-generation requests grounded in real-world.

\textbf{Front-end Web Design.}
For web design, we directly use existing websites as target artifacts. Specifically, we auto-curate publicly accessible and high-quality websites from design award platforms (e.g., \href{https://www.awwwards.com/}{Awwwards}) and product discovery sites (e.g., \href{https://saaslandingpage.com/}{SaaS Landing Page}), and organize them along two dimensions: site type and industry domain. For each website, we capture its rendered layout, navigation structure, and interactive components, and then use LLM-based reverse engineering to convert these observations into a structured design specification for the corresponding task instance.

\textbf{Poster Generation.}
To cover a diverse set of common real-world scenarios, we automatically collect source content associated with practical poster-creation needs, such as data-report visualization, product promotion, event advertising, and social advocacy. The collected materials span multiple domains, including technology, health, environment, and business, and are organized according to these poster-use scenarios. We then use LLMs to transform each source into a poster task specification that defines the target audience, core message, and key content blocks to be presented.

\textbf{Data Visualization.}
Unlike the previous task categories, data visualization does not start from existing artifacts as direct task inputs. Instead, we first survey high-quality visualization examples from open data portals (e.g., \href{https://ourworldindata.org/}{Our World in Data}) and scientific publications to derive a taxonomy of visualization types, subject domains, and analytical goals. Based on this taxonomy, we sample task configurations and prompt LLMs to instantiate them into concrete task specifications. For each resulting specification, we further generate a matching data table that is consistent with the target visualization and analytical objective, enabling end-to-end evaluation of visualization generation.

\textbf{Report Generation.}
For report generation, we build tasks on top of publicly available, real-world tabular datasets from open-data platforms (e.g., \href{https://www.kaggle.com/datasets}{Kaggle}), covering diverse domains such as e-commerce, finance, healthcare, education, and technology. Based on these grounded data sources, we use LLMs to construct report-generation task specifications along several key dimensions, including report type, required sections, and analysis dimensions. Each task specification is explicitly grounded in the underlying dataset by linking the requested analyses and report components to concrete data columns.

For each generated instance, we retain three components: the source package $S$ (including \texttt{source\_brief.md} and, when applicable, associated data files), the structured task specification $T$ (\texttt{task\_input.json}), and the natural-language instruction $I$ (\texttt{instruction.md}). Since these downstream tasks do not admit a single canonical answer, our validation procedure does not rely on reference outputs. Instead, in the third stage, we employ a verifier LLM to assess whether $T$ and $I$ are both information-complete and well-grounded in the source package $S$, and filter out instances that are inconsistent, underspecified, or weakly supported by the source content.
Because the entire pipeline is automated, the benchmark can be continuously refreshed as underlying content sources evolve, allowing it to better keep pace with changing real-world user needs. In the current version, we use Claude-4.6-Opus and GPT-5.2 as the primary generators in the pipeline, resulting in a total of 677 task instances. The distribution of these instances across task categories is shown in Figure~\ref{fig: pop_result}, and the task schemas together with additional case studies are provided in Appendix~\ref{appendix: task_schema}.

\begin{figure*}[htp]
\centering
\begin{minipage}[c]{0.42\textwidth}
    \centering
    \includegraphics[width=\linewidth]{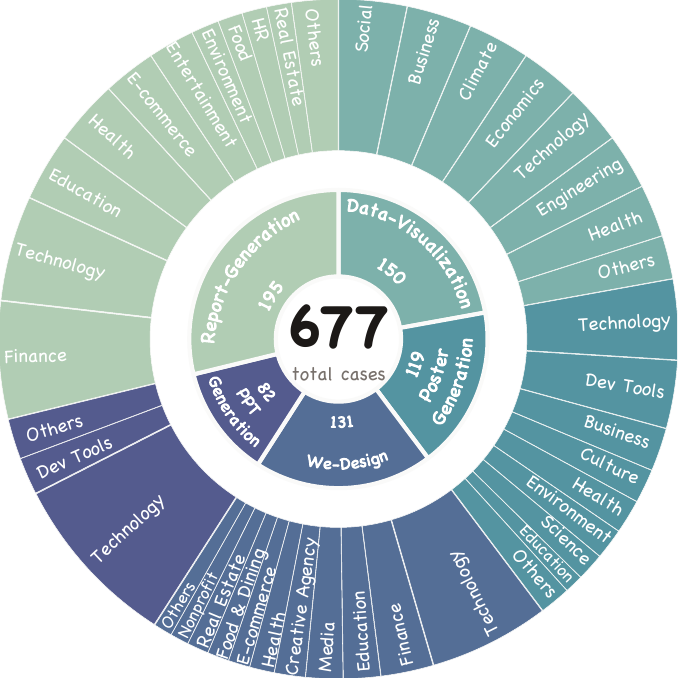}
    \subcaption{Case statistics}
    \label{fig: pop_result}
\end{minipage}\hfill
\begin{minipage}[c]{0.56\textwidth}
    \centering
    \scriptsize
    \setlength{\tabcolsep}{3pt}
    \renewcommand{\arraystretch}{1.0}
    \begin{tabular}{>{\centering\arraybackslash\scriptsize}m{0.18\linewidth} >{\centering\arraybackslash}m{0.04\linewidth} >{\raggedright\arraybackslash}m{0.68\linewidth}}
    \toprule
    \textbf{Task} & \textbf{\#} & \textbf{Skills} \\
    \midrule
    \colorlet{currentSkillColor}{tPres}%
    \textcolor{tPres}{\textbf{Presentation\newline Generation}} & 6 &
    \sk{https://github.com/anthropics/skills/blob/main/skills/pptx/SKILL.md}{anthropics-pptx}, 
    \sk{https://github.com/zarazhangrui/frontend-slides}{frontend-slides}, \sk{https://github.com/MiniMax-AI/skills/tree/main/skills/pptx-generator}{minimax-pptx}, \sk{https://clawic.com/skills/powerpoint-pptx}{powerpoint-pptx}, \sk{https://github.com/hugohe3/ppt-master/tree/main/skills/ppt-master}{ppt-master}, \sk{https://github.com/claude-office-skills/skills/blob/main/pptx-manipulation/SKILL.md}{pptx-manipulation} \\
    \midrule
    \colorlet{currentSkillColor}{tWeb}%
    \textcolor{tWeb}{\textbf{Front-end\newline Web Design}} & 8 &
    \sk{https://github.com/kesslerio/ultimate-frontend-design-openclaw-skill/blob/main/SKILL.md}{frontend-design-ultimate}, \sk{https://github.com/curiositech/some_claude_skills/blob/main/.claude/skills/web-design-expert/CHANGELOG.md}{web-design-expert}, \sk{https://github.com/jordanhubbard/loom/blob/main/personas/default/web-designer/SKILL.md}{loom}, \sk{https://github.com/jezweb/claude-skills/blob/main/plugins/web-design/skills/seo-local-business/SKILL.md}{seo-local-business}, \sk{https://github.com/openclaw/skills/blob/main/skills/mpociot/superdesign/SKILL.md}{SuperDesign}, \sk{https://github.com/mrgoonie/claudekit-skills/blob/main/.claude/skills/ui-styling/SKILL.md}{ui-styling}, \sk{https://github.com/nextlevelbuilder/ui-ux-pro-max-skill/tree/main/.claude/skills/ui-ux-pro-max}{ui-ux-pro-max}, \sk{https://github.com/mrgoonie/claudekit-skills/blob/main/.claude/skills/web-frameworks/SKILL.md}{web-frameworks} \\
    \midrule
    \colorlet{currentSkillColor}{tPos}%
    \textcolor{tPos}{\textbf{Poster\newline Generation}} & 4 &
    \sk{https://skillsmp.com/skills/antvis-infographic-skills-infographic-creator-skill-md}{antv-infographic}, \sk{https://skillsmp.com/skills/wanshuiyin-auto-claude-code-research-in-sleep-skills-paper-poster-skill-md}{paper-poster}, \sk{https://github.com/ComposioHQ/awesome-claude-skills/tree/master/canvas-design}{canvas-design}, \sk{https://github.com/careerhackeralex/visualize}{visualize} \\
    \midrule
    \colorlet{currentSkillColor}{tData}%
    \textcolor{tData}{\textbf{Data\newline Visualization}} & 6 &
    \sk{https://github.com/anthropics/knowledge-work-plugins/blob/main/data/skills/data-visualization/SKILL.md}{anthropics}, \sk{https://github.com/daymade/claude-code-skills/blob/main/suites/daymade-docs/mermaid-tools/SKILL.md}{mermaid-tools}, \sk{https://github.com/mrgoonie/claudekit-skills/blob/main/.claude/skills/mermaidjs-v11/SKILL.md}{mermaidjs}, \sk{https://github.com/caylent/tufte-data-viz}{tufte}, \sk{https://github.com/nicobailon/visual-explainer/blob/main/plugins/visual-explainer/SKILL.md}{visual-explainer}, \sk{https://github.com/careerhackeralex/visualize/blob/main/skills/visualize/SKILL.md}{visualize} \\
    \midrule
    \colorlet{currentSkillColor}{tRep}%
    \textcolor{tRep}{\textbf{Report\newline Generation}} & 6 &
    \sk{https://skills.sh/claude-office-skills/skills/report-generator}{claude-office}, \sk{https://github.com/dkyazzentwatwa/chatgpt-skills/blob/d4bad33523ffc0cd2a7ac9b61d7e28a982786dba/report-generator/SKILL.md}{chatgpt-pdf}, \sk{https://skills.sh/wwwzhouhui/skills_collection/excel-report-generator}{excel-report}, \sk{https://skillsmp.com/skills/jeremylongshore-claude-code-plugins-plus-skills-skills-19-business-automation-report-generator-skill-md}{business-auto}, \sk{https://github.com/guia-matthieu/clawfu-skills/tree/main/skills/automation/report-generator}{clawfu}, \sk{https://github.com/CuriousLearner/devkit/tree/main/skills/report-generator}{devkit} \\
    \bottomrule
    \end{tabular}
    \subcaption{Skills selected for each task (30 total). Names are clickable.}
    \label{tab: skills}
\end{minipage}
\end{figure*}

\subsection{Skills Collection}
\label{subsec: skills collection}

The open-source skill ecosystem is actively maintained and continues to evolve. We therefore comprehensively collect task-relevant skills from multiple public repositories, including \href{https://clawhub.ai/}{clawhub.ai}, \href{https://skills.sh/}{skills.sh}, \href{https://openskills.space/}{openskills.space}, and \href{https://skillsmp.com/}{skillsmp.com}. Because community-contributed skills vary substantially in quality, we do not include every retrieved skill in our evaluation. As shown later in our results (Section~\ref{subsec: skills analysis}), many skill-augmented settings do not outperform the corresponding base agent. We therefore restrict our benchmark to skills with relatively high community adoption, using download counts as a filtering signal under a cost-conscious setting. Following this procedure, we collect a total of 30 skills for evaluation. Detailed statistics and skill information are provided in Table~\ref{tab: skills}.
\footnote{Because these repositories are continuously updated by the community, the collected skill set is time-sensitive and should be understood as a snapshot of the open skill ecosystem at the time of collection.}

\subsection{Automatic Evaluation Pipeline}
\label{subsec: auto-evaluation}

To comprehensively and efficiently evaluate skill-augmented agents, we design an automatic evaluation pipeline from two complementary perspectives: \textbf{trajectory trace analysis} and \textbf{artifact analysis}. The former focuses on the agent's execution process, while the latter assesses the quality of the final task output.
For trajectory trace analysis, we leverage the Agent Trajectory Interchange Format (ATIF)~\cite{Harbor_Framework}, a unified representation that standardizes execution traces across different agent frameworks and enables consistent parsing and analysis across otherwise heterogeneous systems. Building on ATIF, we further introduce an agent-as-judge procedure that first decomposes each skill into a sequence of intended workflow steps, and then compares these steps against the actual agent trajectory at a finer granularity. This process allows us to evaluate whether the agent invokes the skill at the appropriate stage, whether it follows the prescribed workflow, and to what extent the injected skill shapes the overall execution process.

Artifact analysis evaluates whether the final outputs produced by skill-augmented agents achieve the desired quality in realistic application settings. Concretely, for each task category, we design task-specific evaluation criteria to automatically assess output quality. Our metric design is informed by prior work, such as PPTEval~\cite{zheng2025pptagent}, GenEval~\cite{hong2025frabenchufevalunifiedfinegrained}, and WebArena~\cite{zhouwebarena}, while being adapted to the characteristics of each downstream task. Across task categories, we include several shared dimensions, including \emph{completeness}, which measures whether the output satisfies the requirements specified in the task specification, as well as \emph{content quality} and \emph{visual design}, whose exact definitions are adjusted according to the task type.
Beyond these shared criteria, we further design targeted evaluation procedures for tasks that require more specialized assessment. For web design, in addition to visual evaluation based on rendered screenshots, we use agent-based interaction to simulate human clicking and navigation behavior, allowing us to assess functional completeness and collect finer-grained interface states for downstream quality evaluation. For report generation and data visualization, where factual and numerical correctness is particularly important, we explicitly evaluate \emph{data accuracy}. In report generation, this is assessed through code-based analysis of whether the reported values and conclusions are consistent with the underlying data. In data visualization, we combine artifact inspection with trajectory analysis to verify whether the generated charts correctly use and represent the intended data. More detailed evaluation matrix information is shown in Appendix~\ref{appendix: vlm-based juding}.

\section{Experimental Results}

Based on the proposed {OpenSkillEval} framework, we conduct a systematic evaluation of a range of state-of-the-art agent systems together with their corresponding foundation models. Our evaluation includes Claude Code~\cite{claudecode2025} with the Claude 4.6 series~\cite{anthropic2026opus46}, Codex~\cite{openaicodex2025} with the GPT series~\cite{openaigpt53codexcard2026}, Gemini CLI~\cite{geminicli2025} with the Gemini 3.1 Pro~\cite{gemini31pro2026}, Kimi Code CLI~\cite{kimicli2025} with the Kimi K2.6~\cite{kimiteam2025kimik2openagentic} series, as well as adapted Minimax models~\cite{minimaxm27_2026}, DeepSeek V4 Pro~\cite{deepseekai2026deepseekv4} and GLM-5.1~\cite{zeng2026glm} integrated into the Claude Code framework. More detailed experimental settings are provided in Appendix~\ref{appendix: experiment envir}. Using the automatically constructed benchmark, we analyze skill-augmented agents from multiple perspectives, including trajectory trace behavior (Section~\ref{subsec: trajectory analysis}), overall agent performance (Section~\ref{subsec: model comparison}), and differences in skill quality (Section~\ref{subsec: skills analysis}).

\subsection{Trajectory Trace Analysis: How Agents Follow Skills}
\label{subsec: trajectory analysis}
\begin{figure*}[htp]
\centering
\begin{minipage}[c]{0.5\textwidth}
    \centering
    \includegraphics[width=\linewidth]{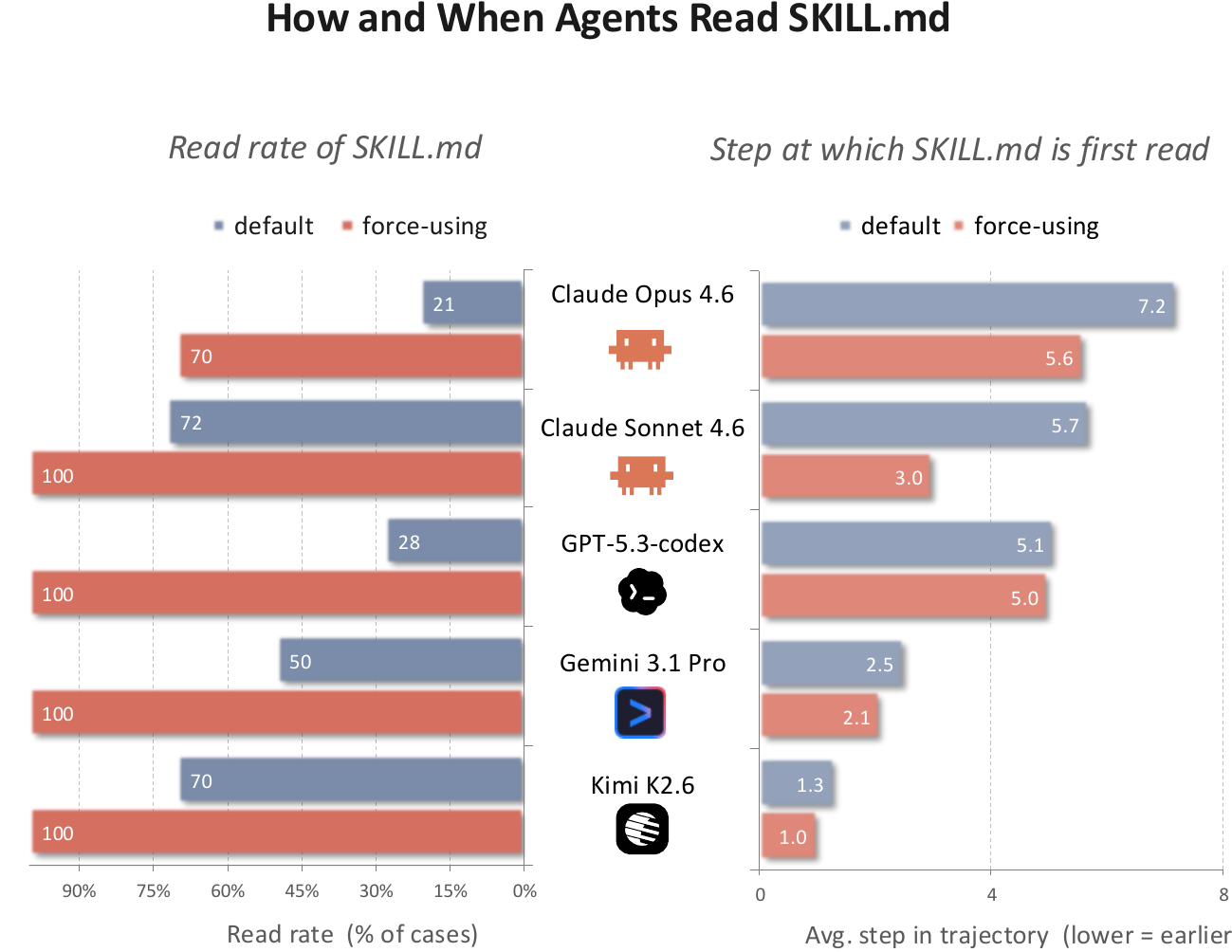}
    \subcaption{}
    \label{fig: trajectory1}
\end{minipage}\hfill
\begin{minipage}[c]{0.49\textwidth}
    \centering
    \includegraphics[width=\linewidth]{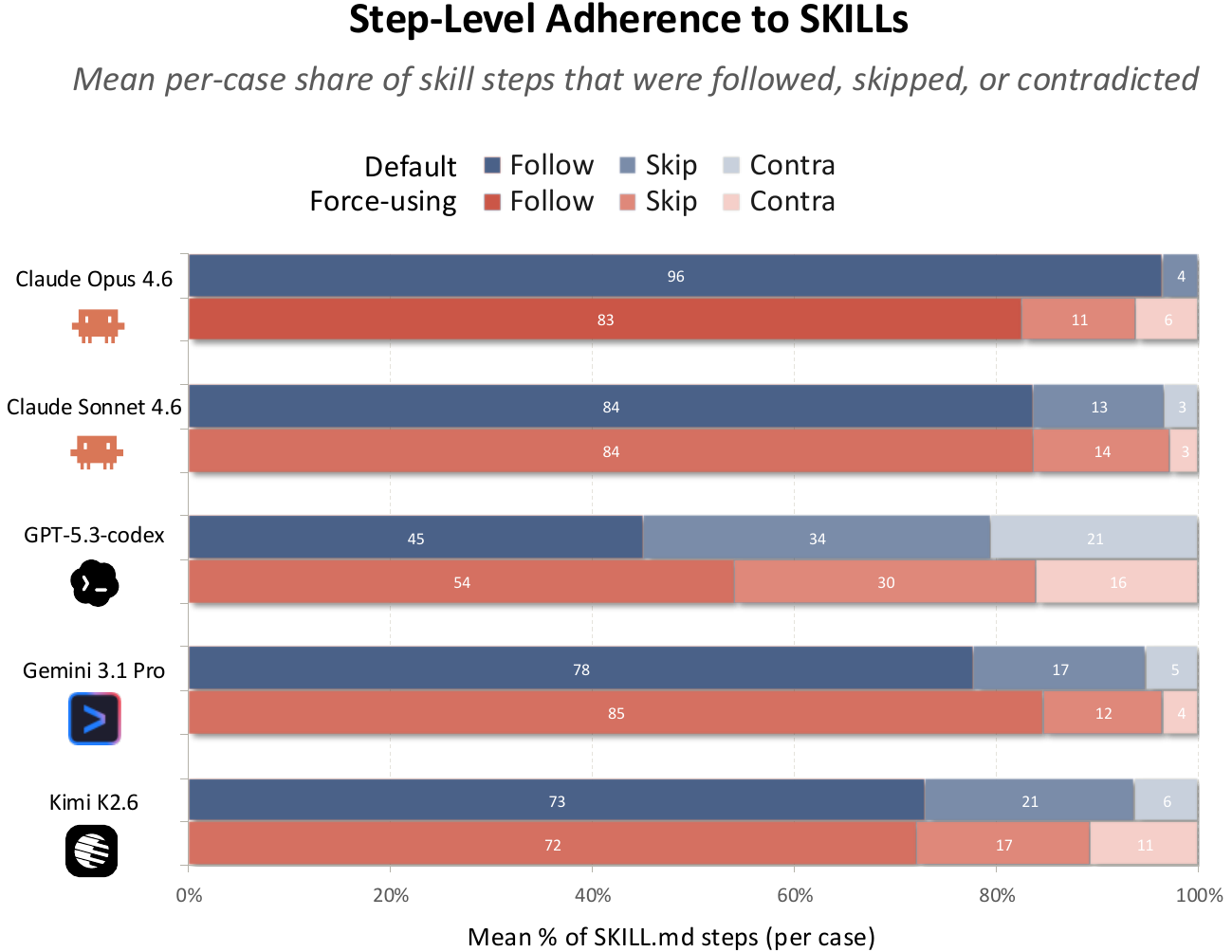}
    \subcaption{}
    \label{fig: trajectory2}
\end{minipage}
\caption{Trajectory-level analysis of how different agent access and follow provided skills. \textbf{(a)} Statistics of \texttt{SKILL.md} access under the default and \emph{force-using} settings, including the proportion of cases in which the skill is explicitly read and the average trajectory step at which it is first accessed. \textbf{(b)} Step-level adherence to skill workflows after skill access, showing the mean proportion of prescribed steps that are followed, skipped, or contradicted across agents under the two settings.}
\vspace{-3mm}
\end{figure*}

Before conducting the large-scale evaluation, we first perform preliminary trajectory-level analyses to better understand how agents actually use provided skills in practice. For this purpose, we place each target skill in the designated initialization path of each CLI-based agent framework, reflecting the common real-world setting in which users download skills into an agent-accessible environment. We then prompt different agents to complete the same downstream tasks under the generated task instructions, and analyze the resulting execution traces to examine whether and how the injected skill is used during task completion.
Our first observation is that, under the default setting, the provided skill often remains effectively unused. As shown in Figure~\ref{fig: trajectory1}, when randomly sampling 100 task instances per agent across task categories, we find that agents explicitly read the corresponding \texttt{skill.md} file in only around 48\% of cases on average. Even for strong models such as Claude Opus 4.6, the read rate is only around 20\%. \textbf{This suggests that simply placing a skill in the accessible environment does not guarantee that the agent will actively discover and use it during execution.}
To enable controlled evaluation of the potential benefits of skill augmentation, we further consider a \emph{force-using} setting that follows the intended skill-usage strategy recommended by agent frameworks. Specifically, we augment the task instruction with an explicit directive to invoke the designated skill. As shown in Figure~\ref{fig: trajectory1}, this intervention substantially increases the probability that agents read and use the provided skill, raising the average read rate to 94\%, and also shifts skill access to earlier stages of the execution trajectory, from an average of 4.4 steps before reading to 3.3 steps.

However, forcing explicit skill usage does not eliminate autonomous agent behavior. By further performing fine-grained trajectory analysis on skills with explicit workflow steps, we find that, as shown in Figure~\ref{fig: trajectory2}, even after agents have explicitly read the provided skills, they still skip prescribed steps (\emph{Skip}) and, in some cases, exhibit behaviors that substantially deviate from the intended procedure (\emph{Contra}). These results indicate that the \emph{force-using} setting can largely mitigate the problem of skills being ignored and can encourage earlier skill access, but agents still retain substantial autonomy in deciding how to execute a task.
This phenomenon is particularly notable for Claude Opus 4.6. Although forcing significantly improves skill access for this model, its overall usage rate still remains noticeably below that of several other agents. A closer analysis reveals substantial variation across task categories: for data visualization and report generation, the skill read rate remains below 50\%, whereas for other, like presentation generation it exceeds 95\%. This pattern suggests a more selective style of skill usage, in which the agent appears to consult skills only when it deems them especially helpful for the task. 
Our trajectory trace analysis further supports this interpretation: before reading the \texttt{skill.md} file, Claude Opus 4.6 tends to spend more steps analyzing the task itself, as reflected by the later step at which \texttt{skill.md} is first accessed in Figure~\ref{fig: trajectory1}. Such selective behavior resembles the more deliberate and rational decision-making patterns previously observed for stronger models in related settings~\cite{ying-etal-2024-intuitive}, such as retrieval-augmented generation. A similar tendency has also been noted by the community~\cite{seleznov2026claudeskills}. \textbf{Taken together, these findings suggest that agents exhibit nontrivial autonomous decision-making in whether and how they use skills.}

\begin{tcolorbox}[
  enhanced, breakable,
  colback=maincolor!6,
  colframe=maincolor!75!black,
  boxrule=0.8pt, arc=4pt,
  left=10pt, right=10pt, top=14pt, bottom=8pt,
  title={Takeaway},
  fonttitle=\bfseries,
  coltitle=white,
  attach boxed title to top left={xshift=10pt, yshift=-\tcboxedtitleheight/2},
  boxed title style={
    colback=maincolor!90!black,
    colframe=maincolor!90!black,
    arc=3pt, outer arc=3pt, boxrule=0pt,
    top=2pt, bottom=2pt, left=8pt, right=8pt,
  },
]
\small
Simply placing a skill in an agent-accessible environment does not guarantee that the agent use it. Even when skill usage is explicitly enforced, agents may still selectively skip or deviate from prescribed workflow steps. This suggests that skill effectiveness in practice depends not only on skill quality, but also on how the agent autonomously chooses to recognize, access, and follow the skill during execution.
\end{tcolorbox}

Since such agent autonomy is itself an important part of how skill augmentation functions in practice, we preserve this behavior in our evaluation. Accordingly, in the subsequent experiments, we evaluate skill augmentation under both \emph{no-skills} and \emph{force-using skills} settings, and assess performance primarily based on the final output artifacts (e.g., screenshots and generated results).

\subsection{Model Comparison: How Different Agents Perform}
\label{subsec: model comparison}

\begin{table*}[htp]
  \centering
  \footnotesize
  \setlength{\tabcolsep}{3.5pt}
  \renewcommand{\arraystretch}{1.20}

  \resizebox{\textwidth}{!}{%
  \begin{tabular}{l*{10}{c}}
    \toprule
    \rowcolor{HeaderBlue}
        \textbf{\textcolor{HeaderText}{Metric}}      &
    \textbf{\textcolor{HeaderText}{Claude Opus 4.6}}      &
    \textbf{\textcolor{HeaderText}{Claude Sonnet 4.6}}     &
    \textbf{\textcolor{HeaderText}{GPT-5.5}}    &
    \textbf{\textcolor{HeaderText}{GPT-5.3-codex}}    &
    \textbf{\textcolor{HeaderText}{GPT-5.2}}    &
    \textbf{\textcolor{HeaderText}{Gemini 3.1 Pro}}    &
    \textbf{\textcolor{HeaderText}{DeepSeek V4}}     &
    \textbf{\textcolor{HeaderText}{GLM-5.1}}     &
    \textbf{\textcolor{HeaderText}{Kimi K2.6}}  &
    \textbf{\textcolor{HeaderText}{MiniMax M2.7}}     \\
    \midrule

    \taskband{\iconPPT}{Presentation Generation}{ppt-generation}
    Content quality   & \cval{4.48}{0.35}  & \cval{4.44}{0.41} & \cval{4.41}{0.39}  & \cval{3.55}{0.52} & \cval{3.74}{0.49} & \cval{3.78}{0.43} & \cval{4.33}{0.33} & \bcval{4.58}{0.29} & \cval{4.14}{0.53} & \cval{4.18}{0.42} \\
    Visual design     & \cval{3.98}{0.30}  & \cval{3.92}{0.27} & \cval{4.01}{0.28}  & \cval{3.04}{0.46} & \cval{3.23}{0.50} & \cval{2.99}{0.64} & \cval{3.79}{0.29} & \bcval{4.09}{0.27} & \cval{3.57}{0.52} & \cval{3.73}{0.36} \\
    Completeness      & \cval{4.68}{0.54}  & \cval{4.63}{0.67} & \cval{4.66}{0.70}  & \cval{4.07}{1.08} & \cval{4.46}{0.74} & \cval{4.24}{0.67} & \cval{4.67}{0.58} & \bcval{4.76}{0.49} & \cval{4.53}{0.75} & \cval{4.44}{0.77} \\
    Fidelity          & \cval{4.51}{0.53}  & \cval{4.31}{0.52} & \bcval{4.88}{0.35} & \cval{4.02}{1.13} & \cval{4.84}{0.38} & \cval{4.61}{0.51} & \cval{4.20}{0.49} & \cval{4.42}{0.54}  & \cval{4.46}{0.72} & \cval{4.18}{0.61} \\
    \rowcolor{StripeRow}
    \textit{Task avg.} & \textit{4.41} & \textit{4.33} & \textbf{\textit{4.49}} & \textit{3.67} & \textit{4.07} & \textit{3.90} & \textit{4.25} & \textit{4.47} & \textit{4.17} & \textit{4.13} \\

    \taskband{\iconPoster}{Poster Generation}{poster-generation}
    Content quality   & \cval{4.56}{0.71}  & \cval{4.37}{0.75} & \bcval{4.60}{0.71} & \cval{4.42}{0.91} & \cval{4.49}{0.76} & \cval{4.43}{0.70} & \cval{4.39}{0.74} & \cval{4.48}{0.74} & \cval{4.49}{0.73} & \cval{4.20}{0.81} \\
    Visual design     & \bcval{4.00}{0.60} & \cval{3.72}{0.77} & \cval{3.82}{0.58}  & \cval{3.19}{0.76} & \cval{3.11}{0.83} & \cval{3.39}{0.84} & \cval{3.55}{0.75} & \cval{3.65}{0.76} & \cval{3.48}{0.84} & \cval{3.10}{0.88} \\
    Completeness      & \bcval{4.12}{0.53} & \cval{3.98}{0.66} & \cval{3.97}{0.63}  & \cval{3.44}{0.79} & \cval{3.43}{0.84} & \cval{3.41}{0.80} & \cval{3.90}{0.65} & \cval{3.95}{0.55} & \cval{3.68}{0.74} & \cval{3.35}{0.85} \\
    \rowcolor{StripeRow}
    \textit{Task avg.} & \bf\textit{4.23} & \textit{4.02} & \textit{4.13} & \textit{3.68} & \textit{3.67} & \textit{3.74} & \textit{3.94} & \textit{4.03} & \textit{3.88} & \textit{3.55} \\

    \taskband{\iconDataViz}{Data Visualization}{data-visualization}
    Insight expression & \bcval{4.40}{0.62} & \cval{4.22}{0.70} & \cval{4.04}{0.77} & \cval{2.79}{1.12} & \cval{3.15}{1.10} & \cval{3.66}{1.00} & \cval{3.96}{0.73} & \cval{4.25}{0.69} & \cval{3.82}{0.85} & \cval{3.31}{1.03} \\
    Visual quality     & \bcval{4.06}{0.62} & \cval{3.99}{0.71} & \cval{3.54}{0.79} & \cval{3.13}{0.86} & \cval{2.92}{0.93} & \cval{3.39}{0.92} & \cval{3.59}{0.72} & \cval{3.81}{0.69} & \cval{3.52}{0.82} & \cval{3.30}{0.91} \\
    Completeness       & \bcval{4.80}{0.42} & \cval{4.67}{0.57} & \cval{4.58}{0.67} & \cval{2.87}{1.35} & \cval{3.46}{1.25} & \cval{4.03}{1.03} & \cval{4.43}{0.71} & \cval{4.69}{0.55} & \cval{4.27}{0.88} & \cval{3.62}{1.13} \\
    Data accuracy      & \bcval{4.98}{0.15} & \cval{4.93}{0.25} & \cval{4.94}{0.26} & \cval{4.25}{1.00} & \cval{4.78}{0.59} & \cval{4.93}{0.27} & \cval{4.95}{0.21} & \cval{4.97}{0.17} & \cval{4.92}{0.30} & \cval{4.82}{0.41} \\
    \rowcolor{StripeRow}
    \textit{Task avg.} & \textbf{\textit{4.56}} & \textit{4.45} & \textit{4.28} & \textit{3.26} & \textit{3.58} & \textit{4.00} & \textit{4.23} & \textit{4.43} & \textit{4.13} & \textit{3.76} \\

    \taskband{\iconReport}{Report Generation}{report-generation}
    Content quality   & \cval{4.20}{0.77}  & \bcval{4.43}{0.71} & \cval{4.03}{0.72}  & \cval{2.53}{0.85} & \cval{3.25}{0.97} & \cval{2.82}{0.71} & \cval{3.90}{0.86} & \cval{3.79}{0.85} & \cval{4.29}{1.00} & \cval{3.35}{0.81} \\
    Visual quality    & \cval{4.60}{0.66}  & \bcval{4.68}{0.67} & \cval{4.52}{0.62}  & \cval{3.81}{0.90} & \cval{3.89}{0.98} & \cval{3.99}{0.71} & \cval{4.38}{0.97} & \cval{4.56}{0.88} & \cval{4.35}{1.09} & \cval{4.11}{0.92} \\
    Completeness      & \cval{4.77}{0.52}  & \bcval{4.81}{0.50} & \cval{4.79}{0.53}  & \cval{3.21}{1.19} & \cval{4.01}{1.14} & \cval{3.14}{0.87} & \cval{4.54}{0.72} & \cval{4.57}{0.71} & \cval{4.62}{1.00} & \cval{4.29}{0.78} \\
    Data accuracy     & \cval{4.87}{0.37}  & \cval{4.74}{0.54}  & \bcval{4.95}{0.24} & \cval{4.66}{0.88} & \cval{4.87}{0.47} & \cval{4.74}{0.60} & \cval{4.76}{0.51} & \cval{4.82}{0.43} & \cval{4.60}{0.98} & \cval{4.50}{0.81} \\
    Fidelity          & \cval{4.58}{0.58}  & \cval{4.44}{0.65}  & \bcval{4.87}{0.36} & \cval{4.46}{0.92} & \cval{4.82}{0.49} & \cval{4.28}{0.89} & \cval{4.21}{0.74} & \cval{4.34}{0.70} & \cval{4.31}{1.00} & \cval{3.89}{0.93} \\
    \rowcolor{StripeRow}
    \textit{Task avg.} & \textit{4.60} & \textit{4.62} & \textbf{\textit{4.63}} & \textit{3.73} & \textit{4.17} & \textit{3.79} & \textit{4.36} & \textit{4.42} & \textit{4.43} & \textit{4.03} \\

    \midrule
    \taskband{\iconWeb}{Front-end Web Design}{web-design}
    Visual design          & \cval{4.62}{0.48}  & \bcval{4.75}{0.42} & \cval{4.71}{0.53}  & \cval{4.12}{0.56} & \cval{4.44}{0.58} & \cval{4.20}{0.84} & \cval{4.56}{0.48} & \cval{4.62}{0.53} & \cval{4.10}{1.30} & \cval{4.43}{0.63} \\
    Responsive             & \cval{4.34}{0.47}  & \cval{4.21}{0.51}  & \bcval{4.56}{0.55} & \cval{4.16}{0.59} & \cval{4.18}{0.60} & \cval{4.06}{0.61} & \cval{4.38}{0.48} & \cval{4.33}{0.48} & \cval{4.02}{1.08} & \cval{4.19}{0.60} \\
    Navigation pass rate   & \cval{4.96}{0.30}  & \cval{4.97}{0.23}  & \cval{4.96}{0.33}  & \cval{4.82}{0.59} & \bcval{4.98}{0.24} & \cval{4.96}{0.26} & \cval{4.96}{0.27} & \cval{4.97}{0.24} & \cval{4.70}{1.13} & \cval{4.93}{0.44} \\
    Interaction pass rate  & \cval{4.92}{0.33}  & \bcval{4.93}{0.33} & \cval{4.91}{0.46}  & \cval{4.73}{0.66} & \cval{4.82}{0.51} & \cval{4.74}{0.68} & \cval{4.91}{0.38} & \cval{4.92}{0.34} & \cval{4.59}{1.15} & \cval{4.84}{0.51} \\
    Data display pass rate & \cval{4.95}{0.22}  & \cval{4.94}{0.21}  & \cval{4.95}{0.32}  & \cval{4.72}{0.71} & \bcval{4.96}{0.14} & \cval{4.93}{0.23} & \cval{4.94}{0.21} & \cval{4.95}{0.18} & \cval{4.70}{1.05} & \cval{4.90}{0.27} \\
    \rowcolor{StripeRow}
    \textit{Task avg.} & \textit{4.74} & \textit{4.75} & \textbf{\textit{4.80}} & \textit{4.47} & \textit{4.66} & \textit{4.55} & \textit{4.73} & \textit{4.74} & \textit{4.40} & \textit{4.63} \\
        \midrule
\rowcolor{HeaderBlue!12}
\rowcolor{HeaderBlue!12}
    \textbf{Overall avg.}
      & \textbf{4.51} & 4.43 & 4.47 & 3.76 & 4.03 & 4.00 & 4.30 & 4.42 & 4.20 & 4.02 \\
    \bottomrule
  \end{tabular}%
  }
  \caption{%
    Per-task evaluation scores by agent (1--5 scale).
    Each cell shows mean$\pm$std across cases.
    \textbf{Bold} marks the best score in each row.
  }
  \label{tab:per-task-scores}
\end{table*}

In this subsection, we compare the performance of different agent systems across downstream tasks under the experimental settings described in Section~\ref{subsec: auto-evaluation}. Across the five task categories, Claude 4.6 within the Claude Code framework and GPT-5.5 within the Codex framework demonstrate the strongest overall performance and the best stability (Table~\ref{tab:per-task-scores}). This suggests that these systems are better able to adapt to different using setting, regardless of whether the injected skills are highly effective or only weakly aligned with the task.
Among them, GPT-Codex does not perform as strongly as its coding-oriented reputation might suggest in our benchmark, which may partly due to weaker agentic capability in more open-ended downstream generation settings. Our skill-level analysis (Section~
\ref{subsec: skills analysis}) further suggests that when the underlying agent capability is limited, even high-quality skills cannot fully realize their potential benefits.
 
 Across tasks, \textbf{1)} Presentation generation and poster generation appear to be the most challenging categories, especially with respect to visual design and layout quality, where average scores are generally below 4. Common failure modes include uneven composition, excessive compression of content, and overlapping elements. As we show later in the skill analysis section (Section~\ref{subsec: skills analysis}), some of these issues can be mitigated when agents are equipped with more detailed, task-specific skills such as \texttt{ppt-master};
\textbf{2)} In contrast, most agents already achieve relatively strong results on front-end web design, especially in terms of interactive functionality, where many systems can already produce reliable usable interfaces. However, there remains a substantial gap between these results and real-world deployment quality. In particular, responsive design scores remain consistently weaker, indicating that adaptation to diverse client-side devices is still inadequate for many generated websites. In addition, visual design issues remain common, including layouts with oversized or undersized elements and unbalanced spacing; 
\textbf{3)} 
For data visualization, most models are already fairly reliable at directly using structured data files to produce charts with high data accuracy. However, their main weakness lies in analytical framing rather than plotting correctness: many generated visualizations fail to clearly surface the most important variable relationships or communicate a strong insight. This is precisely the part of the task where users would most hope carefully designed skills could help, yet current systems still show limited improvement;
 \textbf{4)} 
For report generation, HTML-based rendering often leads to visually polished reports, and most strong models achieve high scores on visual quality. However, content quality and fidelity remain noticeably weaker. In many cases, the generated reports lack a deeper analytical structure: claims are often descriptive rather than rigorously supported, and systematic reasoning steps such as significance testing, robustness checks, or more explicit comparative analysis are frequently missing. This again suggests that current agent systems are better at producing polished artifacts than at carrying out the full analytical process behind them.

\begin{figure*}[htb]
\centering
     \includegraphics[scale=0.42]{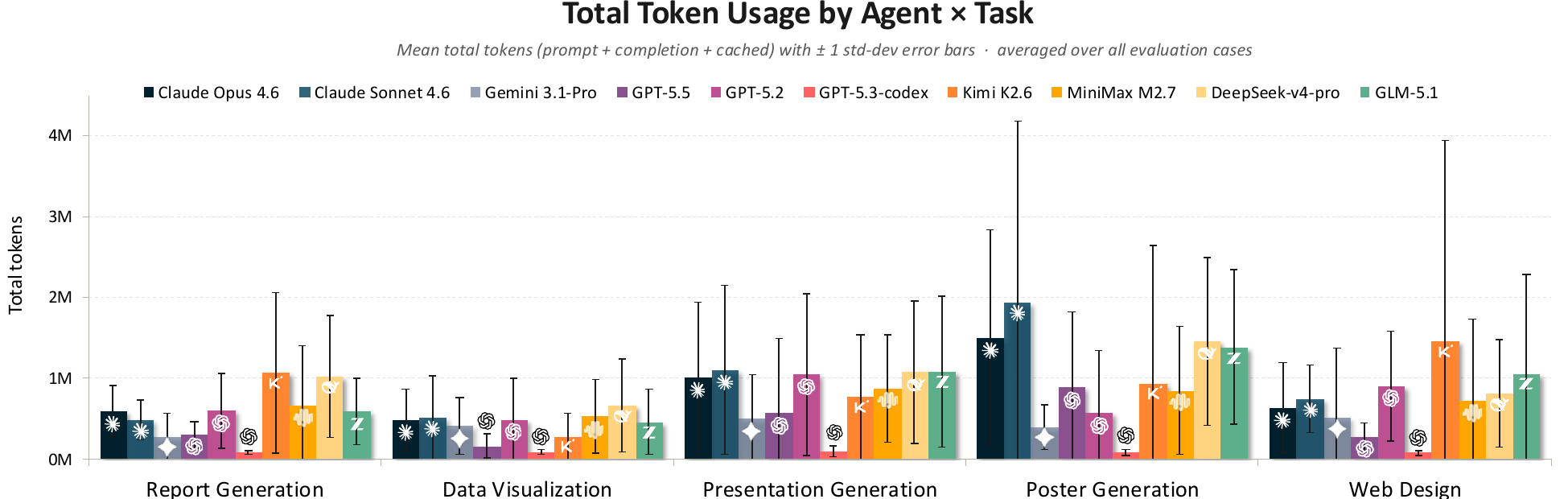}
     \caption{Token usage across agents and tasks.}

     \label{fig: token_usage_all}

\end{figure*}

Beyond task performance, we also analyze token usage across models and tasks, as shown in Figure~\ref{fig: token_usage_all}. We find that the Codex framework consistently exhibits the lowest token consumption across tasks, which is aligned with prior findings~\cite{du2025ockbench} and suggests that the framework is designed with relatively strong efficiency considerations. Notably, GPT-5.5 achieves this efficiency advantage while still maintaining strong overall task performance: compared with agent systems from other model families, it offers a clear efficiency--performance trade-off advantage, and it even consumes fewer tokens than GPT-5.2 within the same series.
Gemini 3.1 Pro follows closely, maintaining comparatively low token usage while also showing relatively strong stability across different skills and task categories. The Claude 4.6 series does not incur excessive token usage overall and remains fairly stable across most tasks. One notable exception is poster generation, where token usage increases substantially; as discussed later in Section~\ref{subsec: skills analysis}, this is partly related to the effect of poster-specific skills. In contrast, the Kimi series shows noticeably weaker stability in agent execution. Besides its larger variance, we also observe frequent cases of abnormally high token consumption. In many instances, Kimi K2.6 enters looping behaviors and continues running until reaching the timeout threshold. Another noteworthy case is GLM-5.1 adapted to Claude Code: because it does not support cached-token mechanisms in the same way as some other systems, its overall execution cost remains comparatively high after adaptation. A more detailed cost breakdown is provided on our project website.
MiniMax shows moderate average consumption overall, but still exhibits relatively large fluctuations across runs.
DeepSeek V4 Pro occupies a favorable middle ground in the cost-performance space, delivering relatively strong performance with moderate token usage, while also offering the practical advantage of open-weight deployment.

Across task categories, poster generation and presentation generation are generally the most token-intensive settings, especially for Claude-family and Kimi-based agents. Our analysis suggests that these tasks often require longer iterative planning, repeated layout adjustment, and multiple rounds of refinement, which together lead to substantially token usage. By contrast, data visualization and report generation are less expensive for most agents, because their workflows are more structurally constrained and their outputs are grounded in clearer data-to-artifact mappings.

\begin{tcolorbox}[
  enhanced, breakable,
  colback=maincolor!6,
  colframe=maincolor!75!black,
  boxrule=0.8pt, arc=4pt,
  left=10pt, right=10pt, top=14pt, bottom=8pt,
  title={Takeaway},
  fonttitle=\bfseries,
  coltitle=white,
  attach boxed title to top left={xshift=10pt, yshift=-\tcboxedtitleheight/2},
  boxed title style={
    colback=maincolor!90!black,
    colframe=maincolor!90!black,
    arc=3pt, outer arc=3pt, boxrule=0pt,
    top=2pt, bottom=2pt, left=8pt, right=8pt,
  },
]
\small
GPT-5.5 and Claude Opus demonstrate the strongest overall performance on these downstream tasks, while GPT-5.5 under the Codex framework also shows a clear efficiency advantage. DeepSeek V4 provides a strong cost--performance compromise together with the practical benefit of open-weight deployment. However, even the best current agents still remain noticeably below the ideal setting in which generated artifacts can be directly used without human intervention, indicating that effective human oversight is still necessary in practice.
\end{tcolorbox}

\subsection{Skill Analysis: How Different Skills Perform}
\label{subsec: skills analysis}

\begin{figure*}[htb]
\centering
     \includegraphics[scale=0.36]{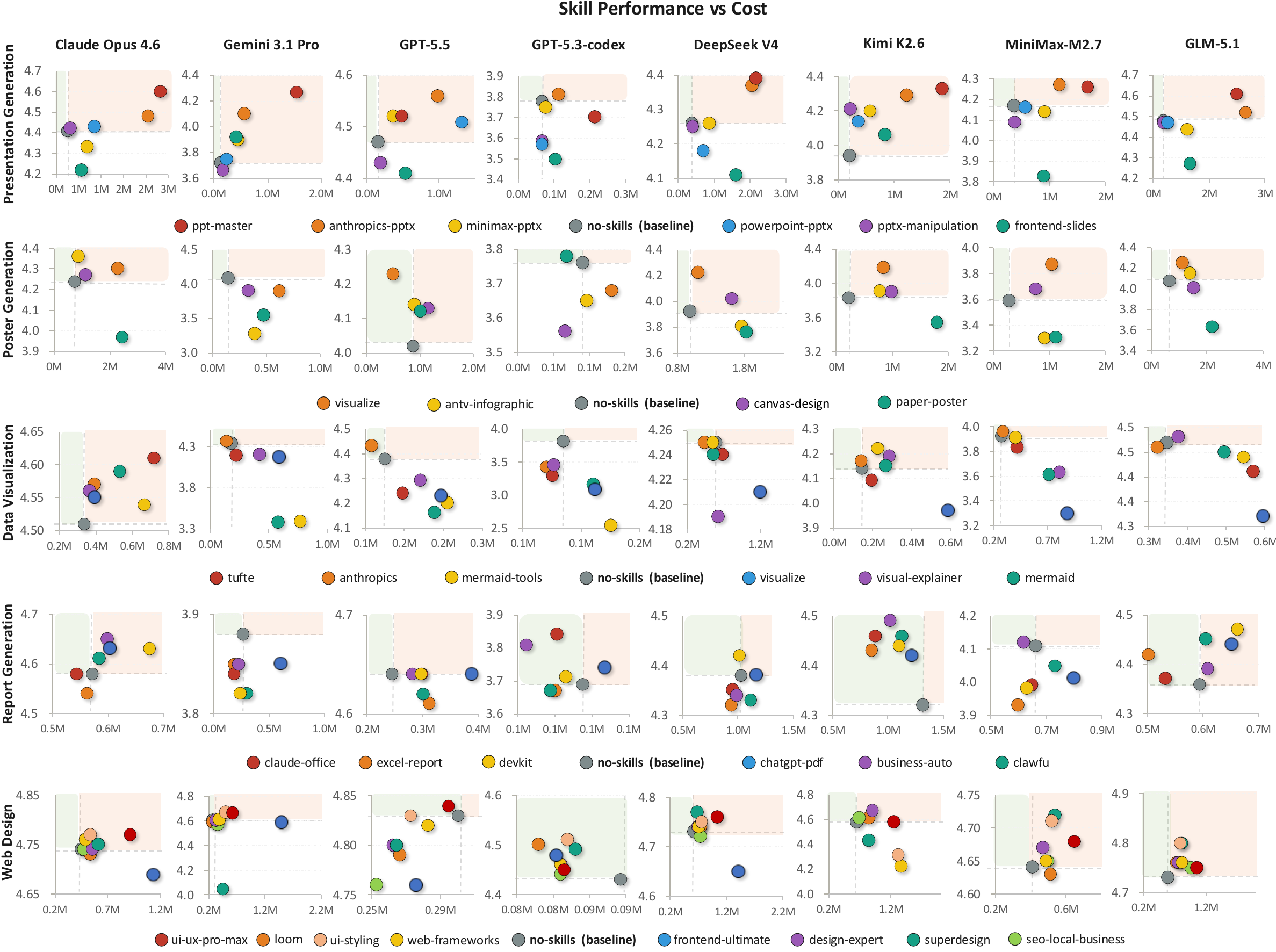}
\caption{Skill performance versus cost across tasks and agent systems. Each subplot corresponds to one model-task pair, where the x-axis shows average token cost and the y-axis shows overall task performance. Colored points denote different skills, while the gray point marks the \emph{no-skills} baseline. The dashed vertical and horizontal lines indicate the baseline cost and performance, respectively, so that points in the upper-left region represent the most desirable outcomes: higher quality at lower cost. The results show that skill augmentation is highly heterogeneous across models and tasks: some skills consistently improve performance, while others increase cost without yielding meaningful gains.}
     \label{fig: skills_main}
     \vspace{-3mm}
\end{figure*}

We compare the effects of different skills across tasks and agent systems. As shown in Figure~\ref{fig: skills_main} (full result is shown in Figure~\ref{fig: skill_pareto_all}), the impact of skill augmentation varies substantially across models, tasks, and individual skills.  For example, GPT-5.3-codex agents generally make weaker use of skills, and in many cases perform worse with skills than without them. This suggests that skill augmentation is not universally beneficial, and that the value of a skill depends not only on the skill itself, but also on how well the target agent can recognize, interpret, and execute it. Nevertheless, some skills provide consistent gains across multiple agents. For instance, \texttt{ppt-master} and \texttt{anthropics-pptx} substantially improve presentation generation, while \texttt{visualize} is particularly effective for poster generation. With the aid of effective skills, even models with relatively modest base performance, such as Kimi K2.6, can improve from around 3.9 to 4.3 in average score, approaching the performance of Claude Opus. From the perspective of inference cost, however, skill augmentation is typically much more expensive. Measured by total token usage (prompt + completion + cache), runs with skills generally consume around 3-5$\times$ more tokens than their no-skills counterparts. As also suggested by Figure~\ref{fig: skills_main}, higher token consumption is often accompanied by better task performance, indicating a clear output-scaling pattern in skill-augmented settings.
To better understand the behavior of skills and the insights they provide, we divide our analysis into two parts. The first focuses on visually dominated artifact-generation tasks, including poster generation, presentation generation, and front-end
web design, where the main effects of skills are reflected in visual language, layout control, and stylistic constraints. The second focuses on reasoning-intensive and analysis-intensive tasks, namely data visualization and report generation, where the primary contribution of skills lies in structuring the problem-solving process and guiding analytical reasoning.

\begin{figure*}[htp]
\centering
\begin{minipage}[c]{0.5\textwidth}
    \centering
    \includegraphics[width=\linewidth]{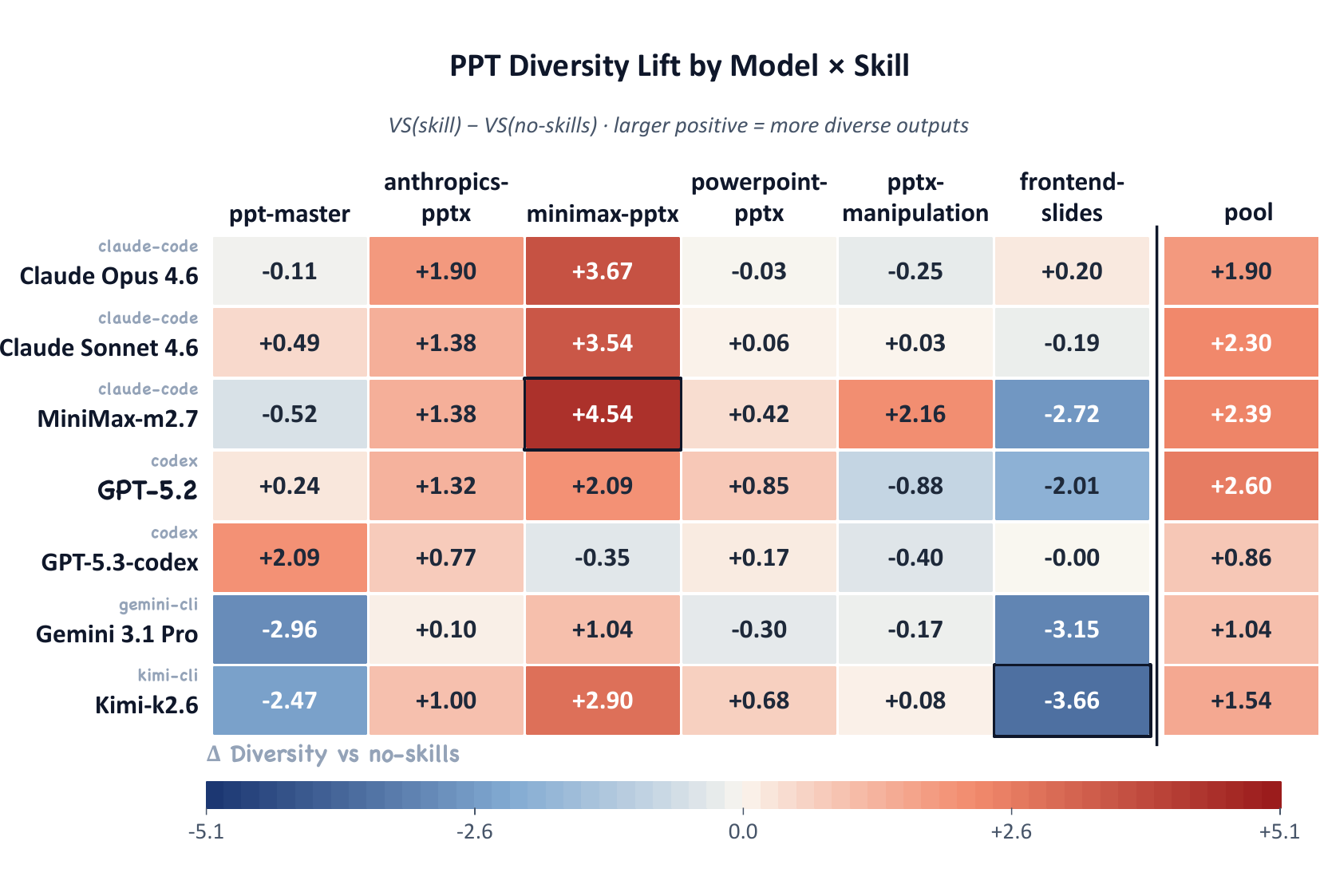}
    \subcaption{}
    \label{fig: impact}
\end{minipage}\hfill
\begin{minipage}[c]{0.49\textwidth}
 \centering
    \includegraphics[width=\linewidth]{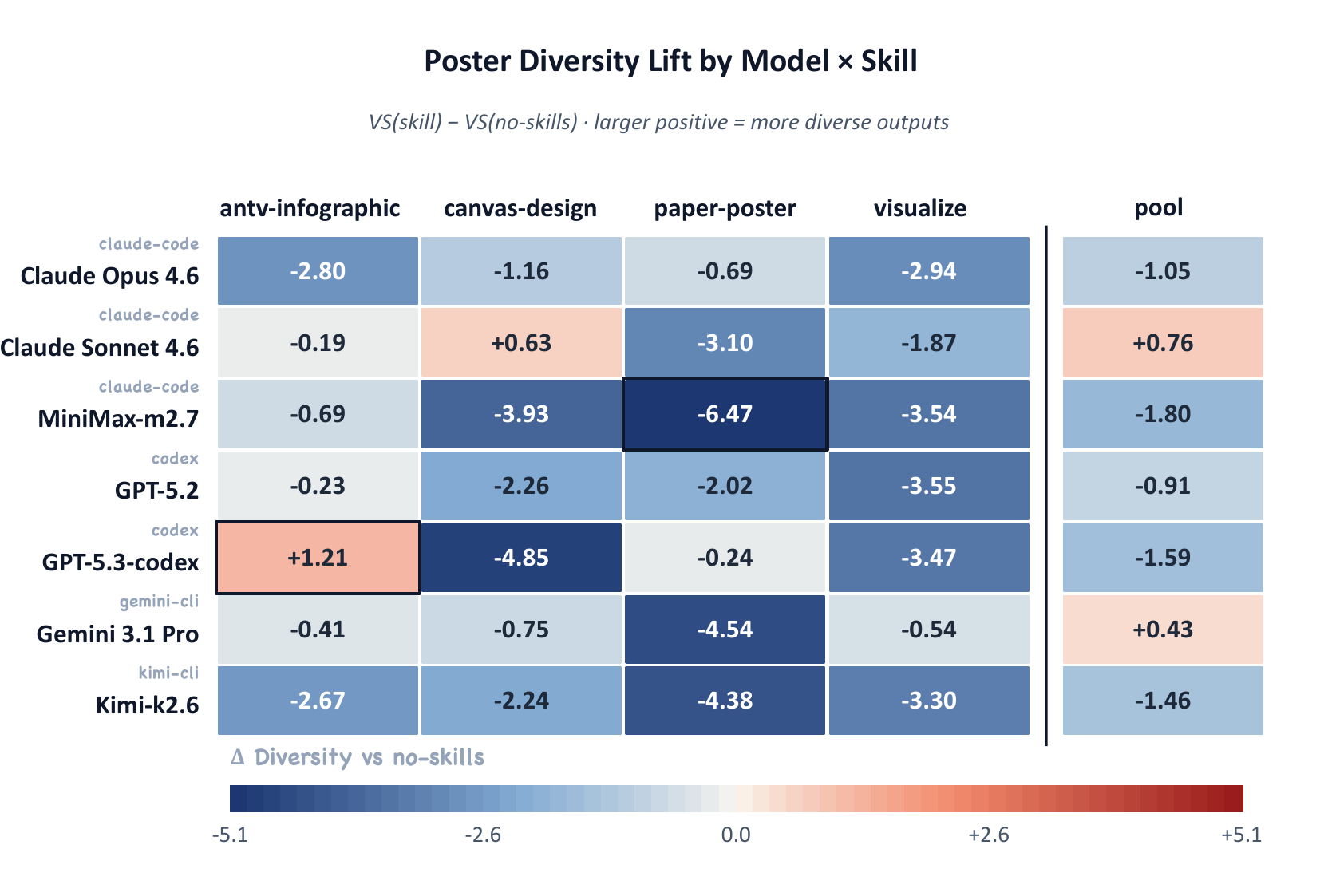}
    \subcaption{}
    \label{fig: impact2}
\end{minipage}
\caption{Impact of skills on stylistic diversity relative to the \emph{no-skills} baseline, measured by changes in within-group Vendi Score computed from CSD-ViT-L style embeddings. Positive values indicate more diverse outputs under a given skill. \textbf{(a)} Presentation generation. \textbf{(b)} Poster generation. The \emph{pool} column aggregates outputs across all skills for cross-skill analysis.}
\end{figure*}

\textbf{Visually Oriented Artifact Generation.}
For visually oriented artifact generation tasks, like presentation generation, poster generation, design language is a central component of the artifact. Therefore, beyond rubric-based quality evaluation, we further examine the stylistic diversity of generated artifacts. To quantify stylistic diversity, we encode outputs using CSD-ViT-L style embeddings~\cite{somepalli2024measuring} and compute the Vendi Score within each skill group, which measures intra-group diversity. To ensure fair comparison, we control for sample size across settings and compare each skill-specific group against a matched \emph{no-skills} baseline, while also reporting a pooled multi-skill condition (\emph{pool}) for cross-skill analysis. Figure~\ref{fig: impact} and Figure~\ref{fig: impact2} summarize the results for presentation and poster generation, respectively. Front-end web design is reported in Figure~\ref{fig: impact3}, because the overall diversity effect of skills is small (less than 0.02), we do not conduct further diversity analysis.
We observe that \textbf{skill usage does not necessarily increase stylistic diversity, even across skills}. For presentation generation, some skills, such as \texttt{anthropics-pptx} and \texttt{minimax-pptx}, consistently increase diversity relative to the \emph{no-skills} baseline, whereas others, such as \texttt{frontend-slides}, substantially reduce it. Most remaining skills induce only modest changes. In contrast, for poster generation, diversity generally decreases under skill augmentation, suggesting that poster-oriented skills tend to impose stronger stylistic constraints on the output. Moreover, the diversity effect of a skill varies substantially across models (with Pearson correlations of 0.52 for presentation generation and 0.28 for poster generation), indicating that, much like overall task performance, stylistic outcomes emerge from the interaction between the skill and the underlying agent rather than from the skill alone.
The pooled multi-skill condition provides an additional perspective on cross-skill variation. For presentation generation, the \emph{pool} condition is consistently more diverse than \emph{no-skills}, indicating that different presentation skills indeed induce meaningfully different visual languages. For poster generation, however, the pooled condition shows little or no diversity advantage over \emph{no-skills}. This suggests that although different poster skills are different, most of them tightly constrain outputs into a small number of fixed visual idioms, so the overall ecosystem still spans only a limited stylistic space.

To better understand how skill design contributes to both output quality and stylistic diversity, we conduct a more fine-grained analysis of the visual-generation skills in Table~\ref{tab:skill-quant}. We characterize each skill along two orthogonal dimensions, corresponding to two complementary forms of experience encoded in the skill. The first is \textbf{design priors}, which capture the reusable visual experience packaged with the skill, including visual template files, reference design documents, design-related data assets, and design-specific content inside \texttt{SKILL.md}. These priors provide concrete design-oriented guidance that informs the model's visual generation behavior. The second is \textbf{procedural constraints}, which capture how strongly the skill prescribes the generation process through explicit directives, such as \texttt{MUST}-style and \texttt{NEVER}-style instructions, together with other structural locks embedded in the workflow. These constraints encode rule-like operational experience that restricts the model's output space and guides it toward more controlled execution.  To operationalize these two dimensions, we first extract a set of countable signals using rule-based heuristics, and then manually map them to coarse-grained 0-5 scores to facilitate trend-level comparison across skills.

\newcolumntype{P}[1]{>{\raggedright\arraybackslash}p{#1}}

\begin{table*}[!htbp]
  \centering
  \footnotesize
  \setlength{\tabcolsep}{4pt}
  \renewcommand{\arraystretch}{1.20}
  \resizebox{\textwidth}{!}{%
    \begin{tabular}{l P{6.6cm} P{3cm} c c c c c c}
    \toprule
    \rowcolor{HeaderBlue}
    \textbf{\textcolor{HeaderText}{Skill}} &
    \textbf{\textcolor{HeaderText}{Key Assets}} &
    \textbf{\textcolor{HeaderText}{Key Constraints}} &
    \textbf{\textcolor{HeaderText}{$A$}} &
    \textbf{\textcolor{HeaderText}{$C$}} &
    \textbf{\textcolor{HeaderText}{Obs.\,$O$}} &
    \textbf{\textcolor{HeaderText}{Obs.\,$D$}} &
    \textbf{\textcolor{HeaderText}{Obs.\,$E$}} &
    \textbf{\textcolor{HeaderText}{Obs.\,Div}} \\
    \midrule

    \taskbandskill{\iconPPT}{Presentation Generation}{ppt-generation}
    ppt-master        & 101 layouts $+$ 33 charts $+$ 640 icons $+$ 39 references
                      & 25 \texttt{MUST} $+$ 10 \texttt{NEVER}
                      & 5 & 5 & \cvalpos{4.27}{} & \cvalpos{3.91}{} & \cvalneg{4.42}{} & \divneg{-0.46} \\
    anthropics-pptx   & 2 references (\texttt{pptxgenjs}, editing) $+$ 78 inline design lines
                      & 1 \texttt{MUST} $+$ 3 \texttt{NEVER}
                      & 3 & 2 & \cvalpos{4.20}{} & \cvalpos{3.65}{} & \cvalneg{4.46}{} & \divpos{+1.12} \\
    minimax-pptx      & 5 references (incl.\ design system) $+$ 8 inline design lines
                      & 9 \texttt{MUST} $+$ 1 \texttt{NEVER}
                      & 2 & 2 & \cvalpos{4.09}{} & \cvalpos{3.47}{} & \cvalneg{4.40}{} & \divpos{+2.49} \\
    powerpoint-pptx   & 3 references (slides, charts, design) $+$ 21 inline design lines
                      & 2 \texttt{MUST}
                      & 1 & 1 & \cvalneg{4.03}{} & \cvalpos{3.25}{} & \cvalneg{4.47}{} & \divpos{+0.27} \\
    frontend-slides   & Viewport-base CSS $+$ 12-style preset library
                      & 13 \texttt{MUST} $+$ 7 \texttt{NEVER}
                      & 2 & 5 & \cvalneg{3.96}{} & \cvalpos{3.52}{} & \cvalneg{4.24}{} & \divneg{-1.65} \\
    pptx-manipulation & Unpack/repack tutorial $+$ 16 inline design lines
                      & ---
                      & 1 & 0 & \cvalneg{4.02}{} & \cvalpos{3.29}{} & \cvalneg{4.43}{} & \divpos{+0.08} \\
    \rowcolor{StripeRow}
    no-skills         & ---
                      & ---
                      & 0 & 0 & \cval{4.04}{} & \cval{3.20}{} & \cval{4.54}{} & --- \\

    \taskbandskill{\iconPoster}{Poster Generation}{poster-generation}
    visualize         & Mandatory skeleton $+$ 9 references $+$ 98 inline design lines
                      & 117 \texttt{MUST} $+$ 28 \texttt{NEVER}
                      & 4 & 5 & \cvalpos{4.05}{} & \cvalpos{3.93}{} & \cvalneg{4.11}{} & \divneg{-2.75} \\
    canvas-design     & 27 inline design lines
                      & 15 \texttt{MUST} $+$ 5 \texttt{NEVER}
                      & 1 & 5 & \cvalneg{3.91}{} & \cvalpos{3.56}{} & \cvalneg{4.09}{} & \divneg{-2.08} \\
    antv-infographic  & External AntV template library
                      & DSL grammar
                      & 1 & 2 & \cvalneg{3.77}{} & \cvalneg{3.47}{} & \cvalneg{3.92}{} & \divneg{-0.83} \\
    paper-poster      & External \LaTeX{} \texttt{tcbposter} class $+$ 256 inline design lines
                      & 21 \texttt{MUST} $+$ 15 \texttt{NEVER}
                      & 2 & 5 & \cvalneg{3.67}{} & \cvalneg{2.88}{} & \cvalneg{4.07}{} & \divneg{-3.07} \\
    \rowcolor{StripeRow}
    no-skills         & ---
                      & ---
                      & 0 & 0 & \cval{3.95}{} & \cval{3.56}{} & \cval{4.14}{} & --- \\
\taskbandskill{\iconWeb}{Front-end Web Design}{web-design}
ui-ux-pro-max      & 16 frontend-stack guides $+$ 14 design CSVs
                   & 44 \texttt{MUST} $+$ 10 \texttt{NEVER}
                   & 5 & 5 & \cvalpos{4.67}{} & \cvalpos{4.41}{} & \cvalneg{4.88}{} & \divneg{-0.01} \\
ui-styling         & 7 references $+$ 2 generator scripts $+$ 321 inline lines
                   & ---
                   & 3 & 0 & \cvalpos{4.66}{} & \cvalpos{4.39}{} & \cvalneg{4.88}{} & \divneg{-0.01} \\
web-design-expert  & 3 references $+$ 220 inline lines
                   & 1 \texttt{NEVER}
                   & 2 & 1 & \cvalpos{4.66}{} & \cvalpos{4.35}{} & \cvalneg{4.90}{} & \divneg{-0.01} \\
loom               & 1 UX-review reference $+$ motivation/personality prompts
                   & 1 \texttt{MUST}
                   & 1 & 1 & \cvalpos{4.65}{} & \cvalpos{4.34}{} & \cvalneg{4.90}{} & \divneg{-0.01} \\
seo-local-business & 3 SEO templates (head, robots, sitemap) $+$ 332 inline lines
                   & 4 \texttt{MUST}
                   & 2 & 2 & \cvalneg{4.64}{} & \cvalneg{4.30}{} & \cvalneg{4.89}{} & \divneg{-0.00} \\
web-frameworks     & 8 references $+$ 2 init scripts
                   & ---
                   & 3 & 0 & \cvalneg{4.63}{} & \cvalneg{4.31}{} & \cvalneg{4.88}{} & \divneg{-0.01} \\
superdesign        & 212 inline design lines, no external references
                   & 2 \texttt{NEVER}
                   & 1 & 1 & \cvalneg{4.55}{} & \cvalneg{4.19}{} & \cvalneg{4.84}{} & \divpos{+0.02} \\
frontend-ultimate  & 3 references $+$ \texttt{site-config} template $+$ 408 inline lines
                   & 1 \texttt{MUST} $+$ 1 \texttt{NEVER}
                   & 2 & 1 & \cvalneg{4.54}{} & \cvalneg{4.24}{} & \cvalneg{4.80}{} & \divpos{+0.01} \\
\rowcolor{StripeRow}
no-skills          & ---
                   & ---
                   & 0 & 0 & \cval{4.65}{} & \cval{4.32}{} & \cval{4.91}{} & --- \\

    \bottomrule
    \end{tabular}%
  }

\caption{Quantifying presentation, poster, and web-design skills along two orthogonal dimensions and relating them to observed quality and stylistic diversity. \textbf{Asset score} ($A$, $0$--$5$) measures the strength of \emph{design priors} packaged with a skill, based on countable signals such as visual template files, reference design documents, design-data size, and design-related line count inside \texttt{SKILL.md}. \textbf{Constraint score} ($C$, $0$--$5$) measures the strength of \emph{procedural constraints}, based on \texttt{MUST}-style and \texttt{NEVER}-style directive counts together with structural locks. \textbf{Obs.\,$O$} denotes the measured overall judge score; \textbf{Obs.\,$D$} denotes the design sub-score, taken directly from the judge for presentation and poster generation and computed as the mean of (Visual Design $+$ Responsive) for web design; \textbf{Obs.\,$E$} denotes the content-oriented effectiveness score, computed as the mean of (Completeness $+$ Fidelity) for presentation generation, the mean of (Content $+$ Completeness) for poster generation, and the mean of three pass-rate for web design; \textbf{Obs.\,Div} denotes the observed visual-diversity score.}

  \label{tab:skill-quant}
\end{table*}

In terms of output quality, our analysis reveals a clear relationship between skill design and observed performance. First, after introducing skills, models almost always achieve higher visual-design scores than in the \emph{no-skills} setting (in all 6 presentation-generation settings, and in roughly half of the poster-generation and front-end web design settings). Moreover, this improvement tends to grow with the richness of the skill's \emph{design priors}. This suggests that well-designed visual priors can directly raise the lower bound of visual artifact quality by providing concrete design guidance that the model can readily exploit.
At the same time, however, content-related metrics such as \emph{Completeness} and \emph{Fidelity} may be slightly degraded. One reason is that skills themselves introduce additional input content and instructions, which the model must process and adapt during generation, thereby placing additional burden on both layout planning and content allocation. 
More importantly, strong visual design alone is not sufficient; a useful skill must also be adaptable across the range of usage scenarios. S. Some skills impose structural choices that are visually beneficial but content-restrictive, which can ultimately reduce overall performance in real-world usage. For example, \texttt{frontend-slides} includes hard density caps on the number of layout elements and strong layout constraints, which makes it visually disciplined but almost inevitably causes content omission when the source material is dense, thereby lowering completeness and overall task scores.
Taken together, these results suggest an \textbf{asset scaling effect in skill design: the richer and better designed the packaged visual priors are, the stronger the overall task performance tends to become.} This pattern is intuitively plausible. In the ideal case, if a skill provides sufficiently broad and high-quality design coverage for the target use cases, then an agent can achieve strong downstream performance simply by executing within that prior space. For instance, with the aid of \texttt{ppt-master}, which covers 22 presentation genres, even models with relatively modest base performance, such as Kimi K2.6 and Gemini 3.1, improve from around 3.9 to 4.3 and from 3.7 to 4.3, respectively, approaching the performance of raw Claude Opus.

In terms of stylistic diversity, although different forms of constraints are not perfectly comparable, we observe a broad negative relationship between constraint strength and output diversity. Skills with stronger procedural constraints tend to produce more visually consistent but less diverse outputs. This pattern is particularly clear in presentation generation: \texttt{frontend-slides} and \texttt{ppt-master}, which have some of the highest constraint scores, also exhibit some of the lowest observed diversity scores. In other words, stronger constraint can improve consistency and sometimes quality, but often at the cost of narrowing the output style space.

\begin{tcolorbox}[
  enhanced, breakable,
  colback=maincolor!6,
  colframe=maincolor!75!black,
  boxrule=0.8pt, arc=4pt,
  left=10pt, right=10pt, top=14pt, bottom=8pt,
  title={Takeaway},
  fonttitle=\bfseries,
  coltitle=white,
  attach boxed title to top left={xshift=10pt, yshift=-\tcboxedtitleheight/2},
  boxed title style={
    colback=maincolor!90!black,
    colframe=maincolor!90!black,
    arc=3pt, outer arc=3pt, boxrule=0pt,
    top=2pt, bottom=2pt, left=8pt, right=8pt,
  },
]
\small
For visually oriented artifact-generation tasks such as presentation generation, skills do not necessarily increase stylistic diversity, but they can substantially improve the model's design quality. In particular, richer and higher-quality \emph{design priors}, such as well-crafted templates, can significantly raise the performance floor of weaker models and also improve stronger models. Meanwhile, stronger \emph{procedural constraints} typically increase output consistency while reducing stylistic diversity.

\end{tcolorbox}

\begin{figure*}[htb]
\centering
     \includegraphics[scale=0.40]{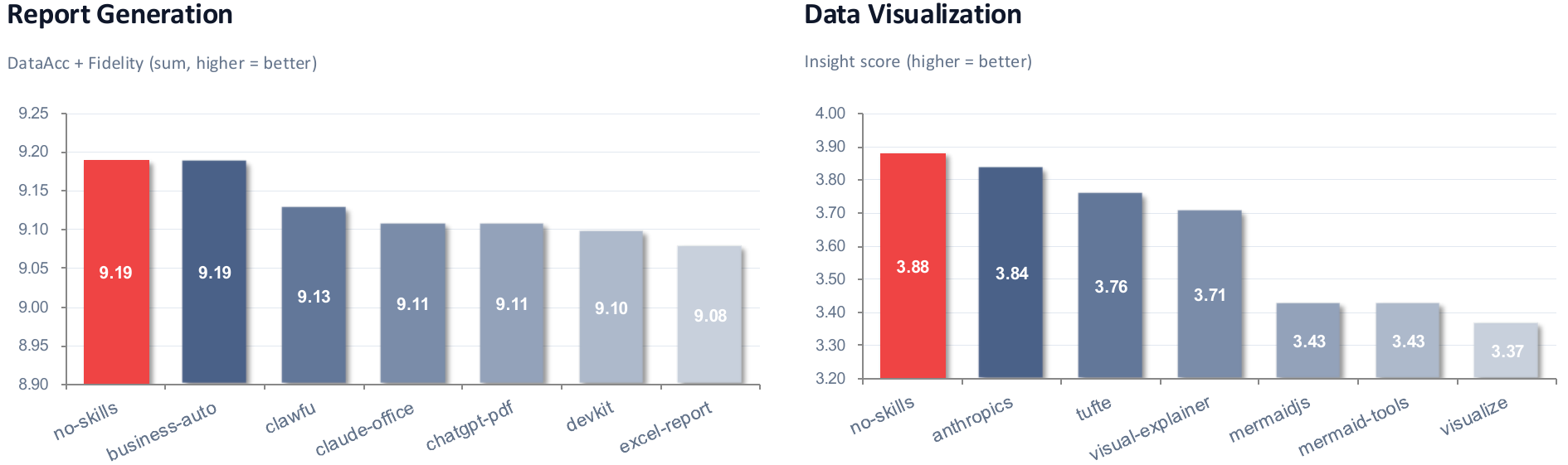}
\caption{Comparison of skill-augmented and \emph{no-skills} settings on reasoning intensive tasks. }

     \label{fig: report_and_data_viz}

\end{figure*}

\textbf{Analytical and Reasoning-Intensive Tasks.}
For more reasoning-intensive tasks, such as report generation and data visualization, require more structured reasoning, more careful data interpretation, and better-designed analysis procedures. Accordingly, beyond overall task scores, we place particular emphasis on reasoning-related indicators. For report generation, we focus on \emph{data accuracy} and \emph{fidelity}, examining whether the model's claims and conclusions are correctly grounded in the underlying data. For data visualization, we place greater emphasis on whether the generated chart clearly communicates the intended insight under the chosen analytical framing.
As shown in Figure~\ref{fig: report_and_data_viz}, we find two main patterns. \textbf{1)} For tasks whose core challenge lies in reasoning and analytical decision-making, existing skills provide only limited benefit. In report generation, most skills perform only on par with, or marginally above, the \emph{no-skills} baseline. The only near-neutral case is \texttt{business-auto}, whose behavior is almost equivalent to \emph{no-skills}, because it contributes nearly no additional analytical guidance. Qualitatively, the generated reports often remain relatively superficial: many claims lack significance testing, deeper validation, or more rigorous experimental support. A similar pattern appears in data visualization. In many cases, agents are able to produce a chart, but fail to clearly surface the underlying relationships among variables or communicate the intended analytical insight. This is also expected given the nature of the current skills: most of them mainly provide tool-level assistance, rather than richer analytical priors, structured workflows, or reasoning-oriented norms. As a result, success on these tasks still depends primarily on the model's own analytical capability; \textbf{2)} From an overall perspective, skill augmentation does not lead to meaningful gains on these tasks: the average improvement in overall score is less than 0.04. One exception is Kimi, where the \emph{no-skills} setting appears worse than several skill-augmented settings. Our trace inspection suggests that this gap is partly attributable to a higher tendency for Kimi to enter looping behavior without skills, rather than to large intrinsic gains from the skills themselves.

\begin{tcolorbox}[
  enhanced, breakable,
  colback=maincolor!6,
  colframe=maincolor!75!black,
  boxrule=0.8pt, arc=4pt,
  left=10pt, right=10pt, top=14pt, bottom=8pt,
  title={Takeaway},
  fonttitle=\bfseries,
  coltitle=white,
  attach boxed title to top left={xshift=10pt, yshift=-\tcboxedtitleheight/2},
  boxed title style={
    colback=maincolor!90!black,
    colframe=maincolor!90!black,
    arc=3pt, outer arc=3pt, boxrule=0pt,
    top=2pt, bottom=2pt, left=8pt, right=8pt,
  },
]
\small
Overall, current reasoning- and analysis-oriented skills provide limited gains, largely because they offer insufficient analytical priors and structured guidance. In practice, their benefits are often marginal, and in some cases the base model without skills may already be the stronger choice.
\end{tcolorbox}

\textbf{Cost Analysis.}
Beyond their impact on artifact quality, we further analyze the cost implications of skill augmentation. As shown in Figure~\ref{fig: skills_main}, in most cases the use of skills does not reduce generation cost relative to the \emph{no-skills} setting. One important reason is that skills themselves introduce additional input content. This effect is especially pronounced for asset-rich skills such as \texttt{ppt-master} and \texttt{visualize}, whose large amount of reference material increases prompt length and leads to a corresponding increase in token consumption. In this sense, the asset scaling effect discussed earlier is accompanied by a parallel cost scaling trend.
Beyond the cost of reading richer skill content, some skills also impose more iterative execution procedures. For example, \texttt{paper-poster} requires the agent to compile LATEX outputs and perform rule-based checking and refinement steps. As a result, certain models, especially the Claude 4.6 series, undergo substantially longer execution loops on this task, often involving more than ten additional refinement steps. This directly contributes to the much larger variance observed in Figure~\ref{fig: token_usage_all}. In extreme cases, the total output-token cost can exceed that of the \emph{no-skills} setting by more than 4$\times$, and for Claude Sonnet the increase can reach nearly 10$\times$. This also indirectly suggests that these models are still not sufficiently strong in these settings, and often require repeated adjustment and refinement before producing satisfactory results.

\section{Human Evaluation} \label{sec: humaneval}

Although our framework is fully automatic and can continuously generate new test cases and conduct evaluation at scale, we further perform human evaluation to assess both the quality of the generated task instances and the reliability of the automatic judgments. Specifically, we involve four senior researchers in natural language processing as human annotators, and randomly sample 100 task instances balanced across the five task categories.
For task-instance quality, annotators score each generated case on three dimensions: \emph{fluency}, \emph{coherence}, and \emph{completeness}, where completeness measures whether the task description is sufficiently clear and well specified for execution. Each dimension is rated on a 1--3 scale. Across all dimensions, the mean human score exceeds 2.98, with an exact agreement rate of 98.8\%. Additional details are provided in Appendix~\ref{appendix: humaneval}. For evaluation quality, given the complexity of the overall evaluation pipeline, we perform human assessment directly on the automatic evaluation results. Annotators are asked to judge whether the assigned score and its accompanying rationale are consistent with the task requirements and the generated output. We use a 1-3 scale: a score of 3 indicates that both the judgment and the rationale are reasonable and well aligned with the task requirements; a score of 2 indicates that the assigned score is reasonable but the rationale is incomplete or not convincing; and a score of 1 indicates that the assigned score itself is unreasonable. Across four annotators, the pooled mean rating is 2.86. Exact four-way agreement accounts for 75.0\% of annotated units, and averagely only 6.01\% of cases receive a rating of 1. These results suggest that the proposed automatic evaluation pipeline is reasonably reliable. In addition, we use another visually strong model from a different family, Gemini 3.1 Pro, as an auxiliary evaluator to further examine the robustness of our automatic evaluation pipeline. On a held-out set of 200 sampled cases, the scores produced by Gemini 3.1 Pro show strong correlation with our primary evaluator, achieving Pearson and Spearman correlations of 0.855 and 0.821, respectively.

\section{Related Work}

\paragraph{Agent skills and procedural augmentation.}
Anthropic recently formalized Agent Skills~\cite{anthropic2025skills}, a \texttt{SKILL.md}-based abstraction for packaging procedural expertise as portable, dynamically loadable artifacts. Building on this paradigm, several works have explored how to construct, discover, and refine skills automatically~\cite{xu2026agentskillslargelanguage, alzubi2026evoskillautomatedskilldiscovery, yang2026autoskillexperiencedrivenlifelonglearning, zheng2025skillweaverwebagentsselfimprove}. As skills proliferate, a parallel line of work has begun to evaluate how effectively models leverage them. SkillsBench~\cite{li2026skillsbenchbenchmarkingagentskills} shows that curated skills can provide substantial but highly variable gains, while self-generated skills offer little benefit on average.  Community efforts such as PinchBench~\cite{pinchbench2026} and WildClawBench~\cite{Ding_WildClawBench} further stress-test agent--skill combinations in realistic workflows. In contrast, our work focuses on downstream application tasks and systematically compares community-contributed skills within the same task settings.
\paragraph{Benchmarks for LLM agents.}
Existing agent benchmarks span software engineering~\cite{jimenezswe}, web
navigation~\cite{zhouwebarena}, and generalist multi-step
reasoning~\cite{liuagentbench}. A persistent limitation is reliance
on static task pools, which saturate quickly and risk contamination~\cite{cao2025generalizableevaluationllmera, NEURIPS2024_1e89c126}. Dynamic
benchmarks mitigate this by releasing fresh problems
on a rolling basis~\cite{tang-etal-2025-evowiki,livebench, jain2024livecodebench}. We extend the dynamic philosophy to the skill-augmented agent
setting, enabling controlled comparison of competing skills under the same task
distribution while jointly measuring effectiveness and efficiency.

\section{Discussion}

The open skill ecosystem for LLM agents is still evolving rapidly. In this work, we build an automatic evaluation framework to make large-scale and continuously updated assessment feasible. However, this design also comes with several limitations. First, due to practical cost constraints, the current version of {OpenSkillEval} does not cover more exhaustive set of available skills. Second, while automation enables scalability and reproducibility, it inevitably abstracts away part of the human-agent interaction process that may matter in real deployment settings.
More broadly, our current evaluation primarily analyzes skill effectiveness from two perspectives: the agent's execution process and the quality of the final output artifact. Although we provide taskaways for skills formalization, we do not directly evaluate skills in isolation. This is mainly because the utility of a skill depends strongly on how different models and agent frameworks interpret and use it. 

\section{Conclusion}

In this work, we present {OpenSkillEval}, an automatic evaluation framework for skill-augmented LLM agents and open-source skills in real-world downstream tasks. By automatically constructing task instances from evolving artifacts, collecting community-contributed skills, and evaluating both agent trajectories and final outputs, {OpenSkillEval} enables a more realistic and scalable analysis of skill effectiveness. Our experiments show that skill availability does not guarantee effective skill use, that the value of skill augmentation depends strongly on the underlying model and agent framework, and that many popular open-source skills provide limited or inconsistent gains. We hope that {OpenSkillEval} can provide practical guidance for selecting both agents and skills in downstream applications, while also offering useful insights for the design, maintenance, and future development of open-source skills.

\bibliography{reference}


\clearpage
\appendix

\section{Technical Appendices and Supplementary Material}

\subsection{Experimental Environment} \label{appendix: experiment envir}

We adopt Harbor~\cite{Harbor_Framework} as the unified execution framework for all experiments, and thank its open-source maintainers for their support and continued development. At runtime, all agents are executed under the same containerized environment based on \texttt{ubuntu:24.04}. To ensure fairness in environment setup and execution, we also provide network access during runtime and use unified timeout settings, with \texttt{build\_timeout\_sec = 1800.0 * 5} and \texttt{timeout\_sec = 900.0 * 5}. All agents are evaluated using the official Responses API service of their corresponding model providers. Unless otherwise specified, all agent frameworks are run with their default parameter settings.\footnote{For runtime stability and reproducibility, we fix the installed versions of the CLI-based agent frameworks in the execution environment.}

For each instance, we provide the agent with the natural-language instruction $I$ (stored as \texttt{Instruction.md}), the structured task specification $T$ (stored as \texttt{task\_input.json}), and any associated source materials or data files required by the task. These additional inputs vary by task category. For example, presentation generation is accompanied by \texttt{source\_brief.md}, which may include tables, figures, and other supporting content; report generation is provided with the corresponding tabular dataset (e.g., \texttt{data.csv}); and data visualization is paired with a structured source file such as \texttt{source\_data.json}.
For each task category, we standardize the expected output format to ensure fair and consistent evaluation across agent systems. Specifically, the required output format is \texttt{.pptx} for presentation generation, a website with \texttt{index.html} as the entry point for front-end web design, \texttt{.png} for poster generation, \texttt{.html} for report generation, and \texttt{.png} for data visualization.

For evaluation, we primarily use Claude Opus 4.6 because of its strong capability, while also using Gemini 3.1 Pro for ablation analysis in Section~\ref{sec: humaneval}. For the agent-as-judge setting, we deploy Claude Opus 4.6 within the Claude Code framework and conduct evaluation in a Docker-based environment.

\subsection{Task-Specific Evaluation Inputs for VLM-Based Judging} \label{appendix: vlm-based juding}

For VLM-based evaluation, we design task-specific input representations and evaluation granularity according to the characteristics of each downstream task.

\paragraph{Presentation Generation.}
For presentation generation, we first convert the generated \texttt{.pptx} files into PDF format using Microsoft PowerPoint (version 16.108.1). We choose this conversion pipeline because we found that commonly used Linux-friendly conversion tools often introduce substantial rendering errors. The resulting PDF is then rasterized page by page into PNG images for per-slide evaluation of \emph{content quality} and \emph{visual design}. For \emph{completeness} and \emph{fidelity}, we provide the evaluator with the full set of slide screenshots, together with the task specification JSON and \texttt{source\_brief.md} (including inline figures), so that it can make a global judgment based on both the output artifact and the source requirements.

\paragraph{Front-end Web Design.}
For front-end web design, we first use Claude 4.6 Opus to simulate user interaction with the generated website through vision- and DOM-grounded browsing. This step is necessary because web evaluation often involves multiple pages, navigation paths, and interaction states that cannot be captured by static screenshot. The browsing agent produces both screenshots and a structured browsing evaluation report containing page-level loading status, discovered sections, missing sections, and navigation outcomes.
Based on these collected screenshots, we evaluate \emph{visual design} and \emph{responsiveness}. For responsiveness, we render the website under multiple viewport settings, including device sizes corresponding to iPhone 13 and iPad 7, to assess layout adaptation across screen sizes. In practice, we find that using only full-page screenshots often leads to excessive image compression and loss of fine-grained details. We therefore adopt a hybrid screenshot strategy: the evaluator receives both full-page screenshots for global structural understanding and viewport-based sequential screenshots (with a viewport height of 900px) for local detail inspection.

\paragraph{Poster Generation and Data Visualization.}
For poster generation and data visualization, the final artifacts are single PNG images, so we directly use the generated images as inputs to the VLM evaluator. The evaluation is grounded in the pre-defined task specification, including criteria such as \emph{coverage of required sections} and \emph{goal insight satisfaction}, in order to assess whether the produced artifact fulfills the requested content and communicative goals.

In addition, for data visualization, we further incorporate an agent-based evaluation component to assess \emph{data accuracy}. Specifically, we provide the evaluator with both the execution trajectory of the task and the original plotting data file (e.g., \texttt{source\_data.json}), and ask it to extract the chart-construction steps and verify whether the generated visualization correctly uses the intended data. This produces a step-level evaluation report focused on the correctness of data usage.

\paragraph{Report Generation.}
For report generation, we first convert the generated HTML report into PDF format, and then apply the same VLM-based screenshot evaluation strategy used for web design: we provide both full-page renderings and segmented screenshots to balance global structure assessment with fine-grained content inspection. For numerical claims and factual consistency, we further introduce an agent-based evaluation procedure. Specifically, we provide the evaluator with the generated HTML report together with the underlying Kaggle dataset, and ask it to analyze the reported claims step by step through code-based verification, producing a step-level evaluation report focused on data consistency and claim correctness.

\begin{table*}[htp]
\centering
\small
\resizebox{\textwidth}{!}{%
\begin{tabular}{@{}llll@{}}
\toprule
\textbf{Task} & \textbf{Dimension} & \textbf{Method} & \textbf{Description} \\
\midrule
\multirow{4}{*}{\textcolor{tPres}{\textbf{Presentation Generation}}}
  & \texttt{content\_quality }      & VLM-judge & Per-slide text quality \\
  & \texttt{visual\_design}       & VLM-judge & Per-slide visual aesthetics \\
  & \texttt{completeness} & VLM-judge & Whole-deck task requirement coverage \\
  & \texttt{fidelity}     & VLM-judge & Whole-deck factual consistency with \texttt{source\_brief.md} \\
\midrule
\multirow{5}{*}{ \textcolor{tWeb}{\textbf{Front-end Web Design}} }
  & \texttt{visual\_design}            & VLM-judge & Aesthetics on full-page screenshots \\
  & \texttt{responsive}                & VLM-judge & Layout consistency across desktop / tablet / mobile \\
  & \texttt{navigation\_pass\_rate}    & Agent     & Playwright tests inter-page navigation links \\
  & \texttt{interaction\_pass\_rate}   & Agent     & Playwright tests interactive components (accordions, modals, etc.) \\
  & \texttt{data\_display\_pass\_rate} & Agent     & Playwright extracts displayed content vs.\ expected items \\
\midrule
\multirow{3}{*}{\textcolor{tPos}{\textbf{Poster Generation}} }
  & \texttt{content\_quality }      & VLM-judge & Data accuracy and traceability to \texttt{source\_brief.md} \\
  & \texttt{visual\_design}       & VLM-judge & Color, layout, typography, polish \\
  & \texttt{completeness} & VLM-judge & Coverage of required sections / metrics \\
\midrule
\multirow{4}{*}{ \textcolor{tData}{\textbf{Data Visualization}}}
  & \texttt{insight\_expression} & VLM-judge & Whether the chart conveys the goal insight \\
  & \texttt{visual\_quality}     & VLM-judge & Color, layout, label completeness \\
  & \texttt{completeness}        & VLM-judge & Task requirement fulfillment \\
  & \texttt{data\_accuracy}      & Agent     & Traces \texttt{trajectory.json} to verify data lineage to \texttt{source\_data.json} \\
\midrule
\multirow{5}{*}{ \textcolor{tRep}{\textbf{Report Generation}}}
  & \texttt{content\_quality} & VLM-judge & Writing quality, clarity, depth of analysis \\
  & \texttt{visual\_quality}    & VLM-judge & Chart selection, color, labels, readability \\
  & \texttt{completeness}     & VLM-judge & Coverage of required sections / KPIs \\
  & \texttt{data\_accuracy}   & Agent     & Python code compares numbers in report vs.\ \texttt{data.csv} \\
  & \texttt{fidelity}         & Agent     &  Verifies extracted claims against \texttt{data.csv}  \\
\bottomrule
\end{tabular}}
\caption{Evaluation matrix across the five \textsc{OpenSkillEval} task categories. Each task is decomposed into multiple evaluation dimensions, which are scored either by a VLM judge (using visual or textual rubrics on a 1--5 scale) or by an evaluation agent (using programmatic verification to produce a score or pass rate). To enable consistent comparison across metrics, pass rates are linearly mapped to the 1--5 scale via $4x + 1$.}
\label{tab:eval-matrix}
\end{table*}

\subsection{Task Input Schemas}
\label{appendix: task_schema}

This section presents the task input schemas used in our benchmark construction pipeline. During automatic case generation, each collected source is transformed into a task-specific structured specification (stored as \texttt{task\_input.json}) following the corresponding schema. These schemas define the core information required for each downstream task, and serve as the intermediate representation from source materials to executable task instances.
\begin{schemabox}{1.\ Data Visualization }
\begin{lstlisting}[style=schemaJson]
{
  // Meta -- required
  "application": "data-visualization",
  "case_id":     "case-climate-trends",
  "language":    "en",                      

  // Style -- optional (omit to test agent autonomy)
  "style": {
    "theme":    "scientific",
    "audience": "researchers and policy makers",
    "tone":     "clean, publication-ready"
  },

  // Goal -- required (one insight per case; chart_type chosen by agent)
  "goal": {
    "id":      "warming-trend",
    "insight": "Show the accelerating global warming trend over 75 years"
  }
}
\end{lstlisting}
\end{schemabox}

\begin{schemabox}{2.\ Poster Generation}
\begin{lstlisting}[style=schemaJson]
{
  // Meta -- required
  "application": "poster-generation",
  "case_id":     "case-01-data-report",
  "language":    "en",

  // Poster constraints -- optional
  "poster": {
    "aspect_ratio": "landscape",   // landscape | portrait | square | A0-landscape | ...
    "audience":     "data-report",
    "tone":         "data-forward, professional",
  },

  // Content brief -- optional
  "brief": {
    "title":     "2025 AI Agent Trends",
    "one_liner": "One-line overview",
    "goal":      "What the audience should take away"
  },

  // Sections -- optional (agent plans structure if omitted)
  "sections": [
    { "id": "overview", "title": "Market Overview",
      "objective": "Show overall market size and growth" }
  ],

  // Metrics -- optional
  "metrics": [
    { "name": "Market Size", "current": "$12B", "target": "$45B (2027)" }
  ]
}
\end{lstlisting}
\end{schemabox}

\begin{schemabox}{3.\ Presentation Generation }
\begin{lstlisting}[style=schemaJson]
{
  // Meta -- required
  "application": "ppt-generation",
  "case_id":     "case-01-internal-review",
  "language":    "en",

  // Deck constraints -- optional
  "deck": {
    "aspect_ratio": "16:9",                  // default 16:9
    "slide_count":  6,                        // omit to let agent decide
    "audience":     "internal product review",
    "tone":         "professional, concise"
  },

  // Content brief -- optional
  "brief": {
    "title":     "SkillArena Q4 Proposal",
    "one_liner": "One-line overview",
    "goal":      "What the audience should understand or decide"
  },

  // Slide blueprint -- optional (agent plans deck if omitted)
  "slides": [
    { "id": "why-now", "title": "Why Now",
      "objective": "Help reviewers understand the launch context" }
    // ...
  ]
}
\end{lstlisting}
\end{schemabox}

\begin{schemabox}{4.\ Report Generation}
\begin{lstlisting}[style=schemaJson]
{
  // Meta -- required
  "application": "report-generation",
  "case_id":     "case-01-sales-analysis",
  "language":    "en",

  // Report constraints -- optional
  "report": {
    "type":     "sales-report",
    "audience": "management",
    "tone":     "professional, data-forward"
  },

  // Content brief -- optional
  "brief": {
    "title":     "2024 Q4 Sales Performance Report",
    "one_liner": "One-line overview",
    "goal":      "What readers should take away"
  },

  // Section blueprint -- optional
  "required_sections": [
    { "section_id": "executive-summary",
      "title":      "Executive Summary",
      "objective":  "Key findings and core metrics at a glance" }
    // ...
  ],

  // KPI definitions -- optional
  "kpis": [
    { "name": "Total Sales", "description": "All Q4 sales revenue" }
  ],

  // Analysis dimensions -- optional
  "analysis_dimensions": ["product", "region", "time_trend"]
}
\end{lstlisting}
\end{schemabox}

\begin{schemabox}{5.\ Web Design }
\begin{lstlisting}[style=schemaJson]
{
  // Meta -- required
  "application": "web-design",
  "case_id":     "case-01-landing-page",
  "language":    "en",

  // Site constraints -- optional
  "site": {
    "type":       "landing-page",
    "page_count": 2,                         // omit to let agent decide
    "audience":   "developers and technical decision-makers",
    "tone":       "modern, professional, bold",
    "responsive": true,                       // default true
    "dark_mode":  false                       // default false
  },

  // Content brief -- optional
  "brief": {
    "title":     "TechPulse -- API Platform",
    "one_liner": "Next-gen API management and deployment",
    "goal":      "Help visitors grasp the value prop and sign up"
  },

  // Page blueprint -- optional
  "pages": [
    { "page_id": "home", "file": "index.html",
      "sections":  ["hero", "features", "testimonials", "cta", "footer"],
      "objective": "Homepage with value prop and sign-up CTA" },
    { "page_id": "pricing", "file": "pricing.html",
      "sections":  ["pricing-tiers", "faq", "cta", "footer"],
      "objective": "Pricing page with three-tier comparison" }
  ],

  // Cross-page navigation -- optional
  "navigation": [
    { "from": "home", "trigger": "navbar 'Pricing' link", "to": "pricing" },
    { "from": "home", "trigger": "'Get Started' CTA button", "to": "pricing" }
  ],

  // In-page interactions -- optional
  "interactions": [
    { "id": "faq-accordion", "page": "home",
      "trigger":  "click on FAQ question",
      "expected": "expand to show the answer; click again to collapse",
      "type":     "toggle" },
    { "id": "pricing-switch", "page": "pricing",
      "trigger":  "toggle 'Monthly / Yearly' switch",
      "expected": "all three tier prices update in sync",
      "type":     "state-change" }
  ],

  // Required data display -- optional
  "data_display": [
    { "id": "pricing-table", "page": "pricing", "type": "table",
      "source_ref":       "comparison table in source_brief.md",
      "expected_content": ["Free", "Pro", "Enterprise", "10K/mo", "1M/mo"],
      "description":      "Feature comparison across three pricing tiers" },
    { "id": "hero-stats", "page": "home", "type": "stats",
      "source_ref":       "core metrics in source_brief.md",
      "expected_content": ["50,000+", "99.99%", "50ms"],
      "description":      "Show 3-4 key numerical metrics in the hero area" }
  ]
}
\end{lstlisting}
\end{schemabox}

\subsection{Human Evaluation}
\label{appendix: humaneval}

\begin{figure*}[htp]
\centering
     \includegraphics[scale=0.65]{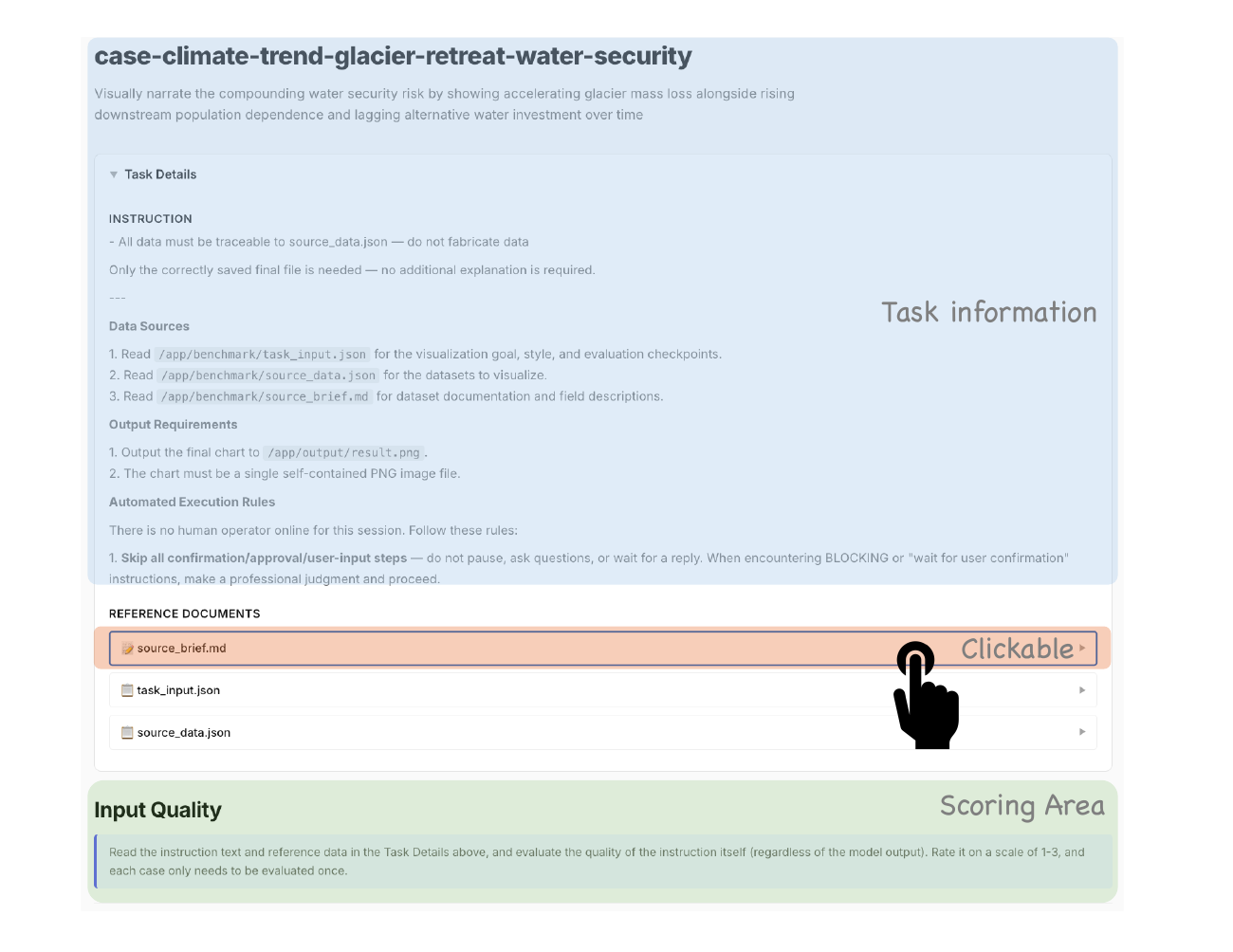}
     \caption{Web-based interface for human evaluation of generated task instances.}
     \label{fig: huamenval_demo1}
\end{figure*}

\begin{figure*}[htp]
\centering
     \includegraphics[scale=0.6]{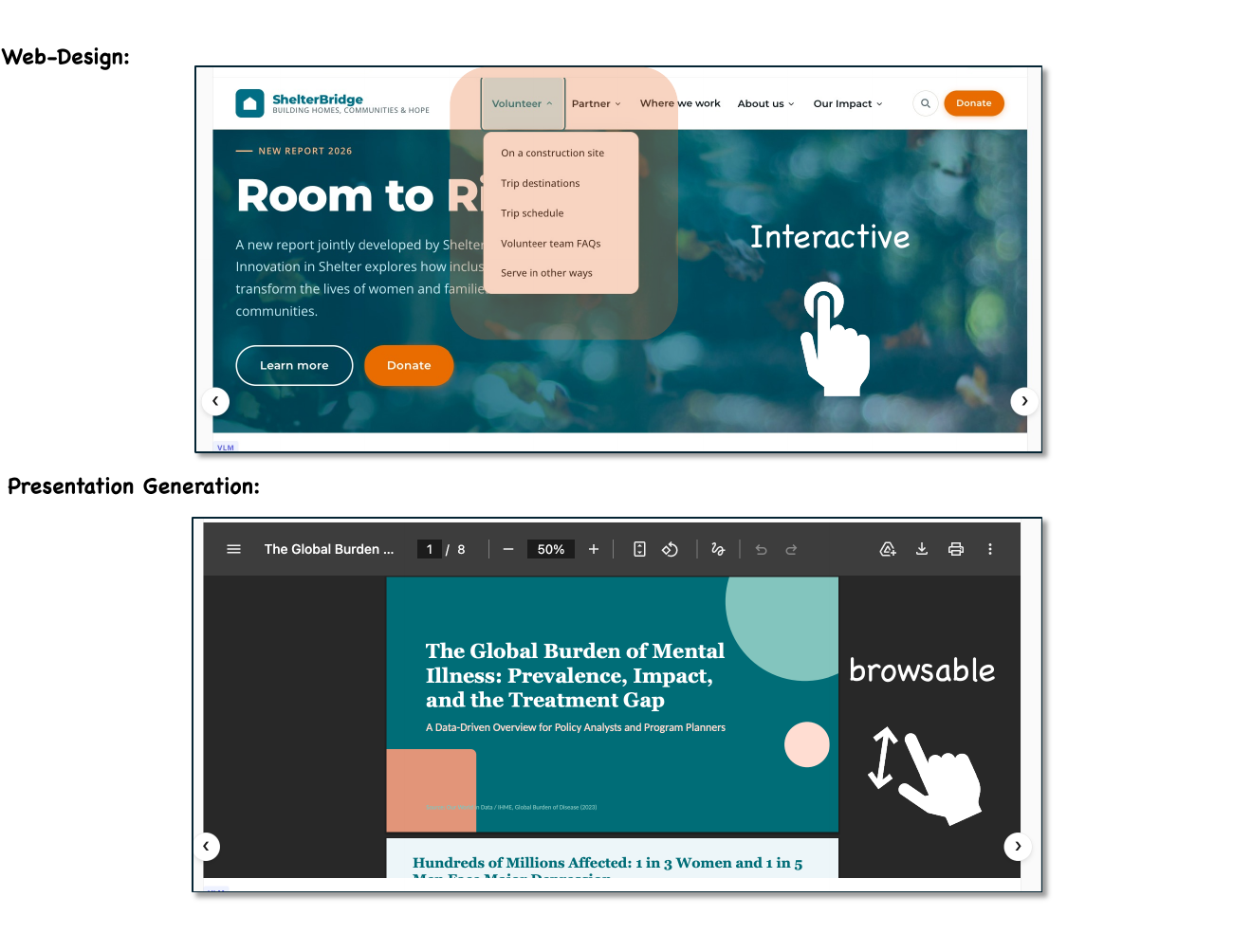}
     \caption{Artifact inspection interface used in human evaluation. The system provides task-specific visualization of generated outputs, including interactive web pages for front-end design tasks and converted PDF views for presentation outputs.}
     \label{fig: huamenval_demo2}
\end{figure*}

To validate both the quality of the automatically generated task instances and the reliability of our Automatic Evaluation Pipeline, we conduct a human evaluation study. Specifically, we randomly sample 100 task instances, balanced across the five task categories, and ask four senior researchers in natural language processing to perform the assessment.
For each sampled instance, evaluators are provided with: (1) the task input, including the natural-language instruction and corresponding task specification; and (2) the generated output artifact. Because many of the task instances involve complex files and multimodal outputs, we build a dedicated web-based interface to support side-by-side inspection. As shown in Figure~\ref{fig: huamenval_demo1}, the interface presents the task details and reference materials in a structured format for easier reading. We also provide task-specific artifact visualizations, as shown in Figure~\ref{fig: huamenval_demo2}. For example, web-design outputs are deployed in an interactive front-end environment for direct inspection, while generated slide decks are converted into PDFs for convenient review.

The detailed guidelines for evaluating the generated task instances are shown below. For artifact evaluation, we directly adopt the evaluation prompts used in our VLM-based automatic evaluation pipeline; full details are provided in Appendix~\ref{appendix: evaluation prompt}.

\begin{tcolorbox}[
  enhanced, breakable,
  colback=white, colframe=maincolor!70,
  boxrule=0.8pt, arc=4pt,
  left=10pt, right=10pt, top=14pt, bottom=8pt,
  title={Task Instances Evaluation Guideline},
  fonttitle=\bfseries,
  coltitle=white,
  attach boxed title to top left={xshift=10pt, yshift=-\tcboxedtitleheight/2},
  boxed title style={
    colback=maincolor, colframe=maincolor,
    arc=3pt, outer arc=3pt, boxrule=0pt,
    top=2pt, bottom=2pt, left=8pt, right=8pt,
  },
]

\small
Read the instruction text and reference data in the task details above, and evaluate the quality of the instruction itself (regardless of the model output). Rate it on a scale of 1--3. Each case only needs to be evaluated once.

\begin{scorebox}{Fluency --- Linguistic quality of the instruction text}
\textbf{3} \;\; Grammatically correct, fluent and natural, accurate terminology \\
\textbf{2} \;\; Minor grammar or wording issues, slightly affects readability \\
\textbf{1} \;\; Multiple grammar errors or unclear wording, severely impedes comprehension
\end{scorebox}

\begin{scorebox}{Coherence --- Logical clarity of the instruction}
\textbf{3} \;\; Clear goal, logically consistent requirements; expected output is unambiguous after reading \\
\textbf{2} \;\; Goal is mostly clear, but some requirements are vague or ambiguous \\
\textbf{1} \;\; Description is chaotic, requirements conflict or are severely vague
\end{scorebox}

\begin{scorebox}{Completeness --- Whether information suffices to complete the task}
\textbf{3} \;\; All essential information present: data source specified, output format clear, constraints stated \\
\textbf{2} \;\; Core information present, but some details are missing; agent must make assumptions \\
\textbf{1} \;\; Critical information is missing; agent cannot reliably complete the task
\end{scorebox}

\end{tcolorbox}

\subsection{More Experimental Result}

\begin{figure*}[htb]
\centering
     \includegraphics[scale=0.32]{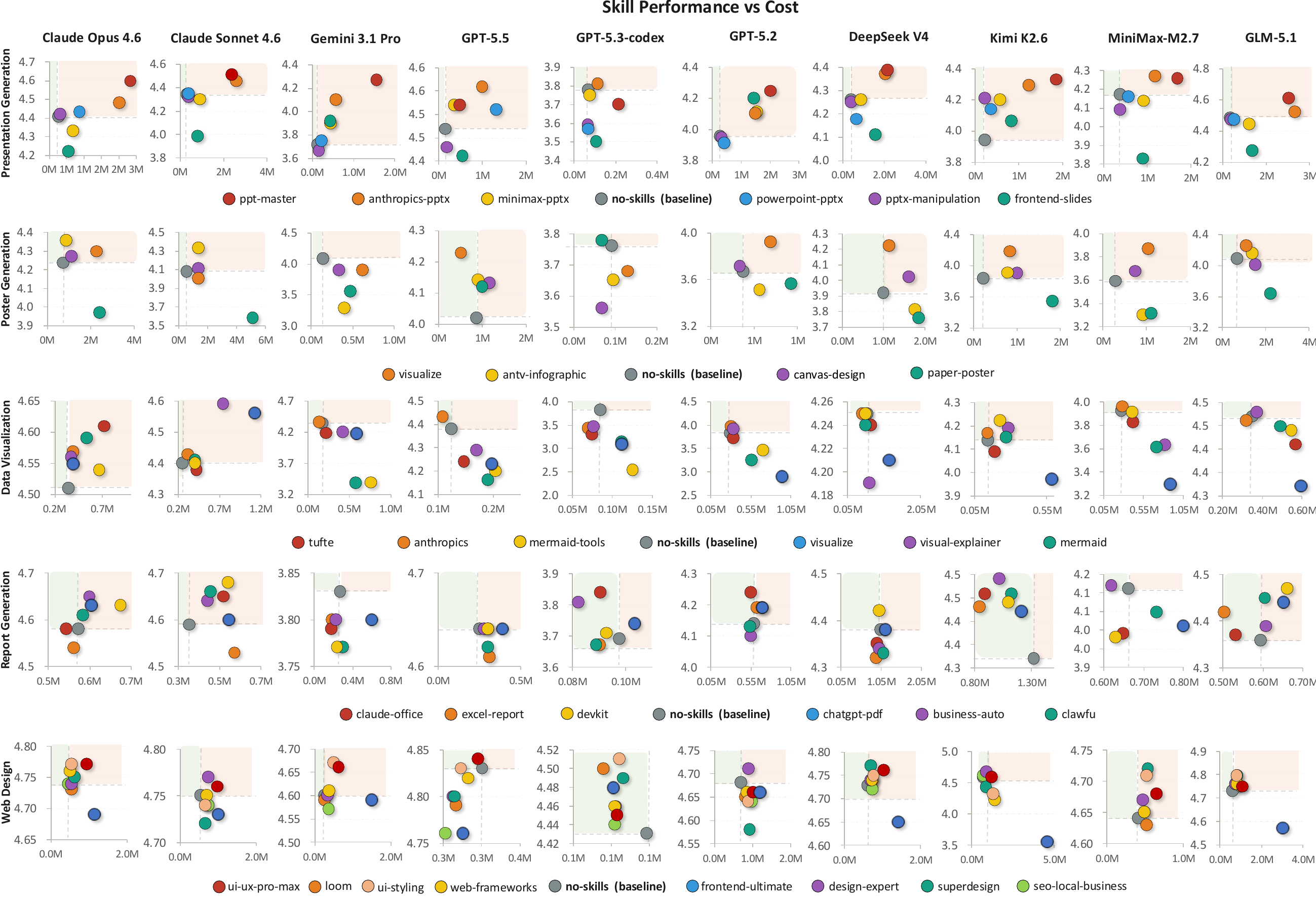}
\caption{Skill performance versus cost across tasks and agent systems. Each subplot corresponds to one model-task pair, where the x-axis shows average token cost and the y-axis shows overall task performance. Colored points denote different skills, while the gray point marks the \emph{no-skills} baseline. The dashed vertical and horizontal lines indicate the baseline cost and performance, respectively, so that points in the upper-left region represent the most desirable outcomes: higher quality at lower cost. The results show that skill augmentation is highly heterogeneous across models and tasks: some skills consistently improve performance, while others increase cost without yielding meaningful gains.}
     \label{fig: skill_pareto_all}

\end{figure*}

\begin{figure*}[htp]
\centering
    \includegraphics[scale=0.35]{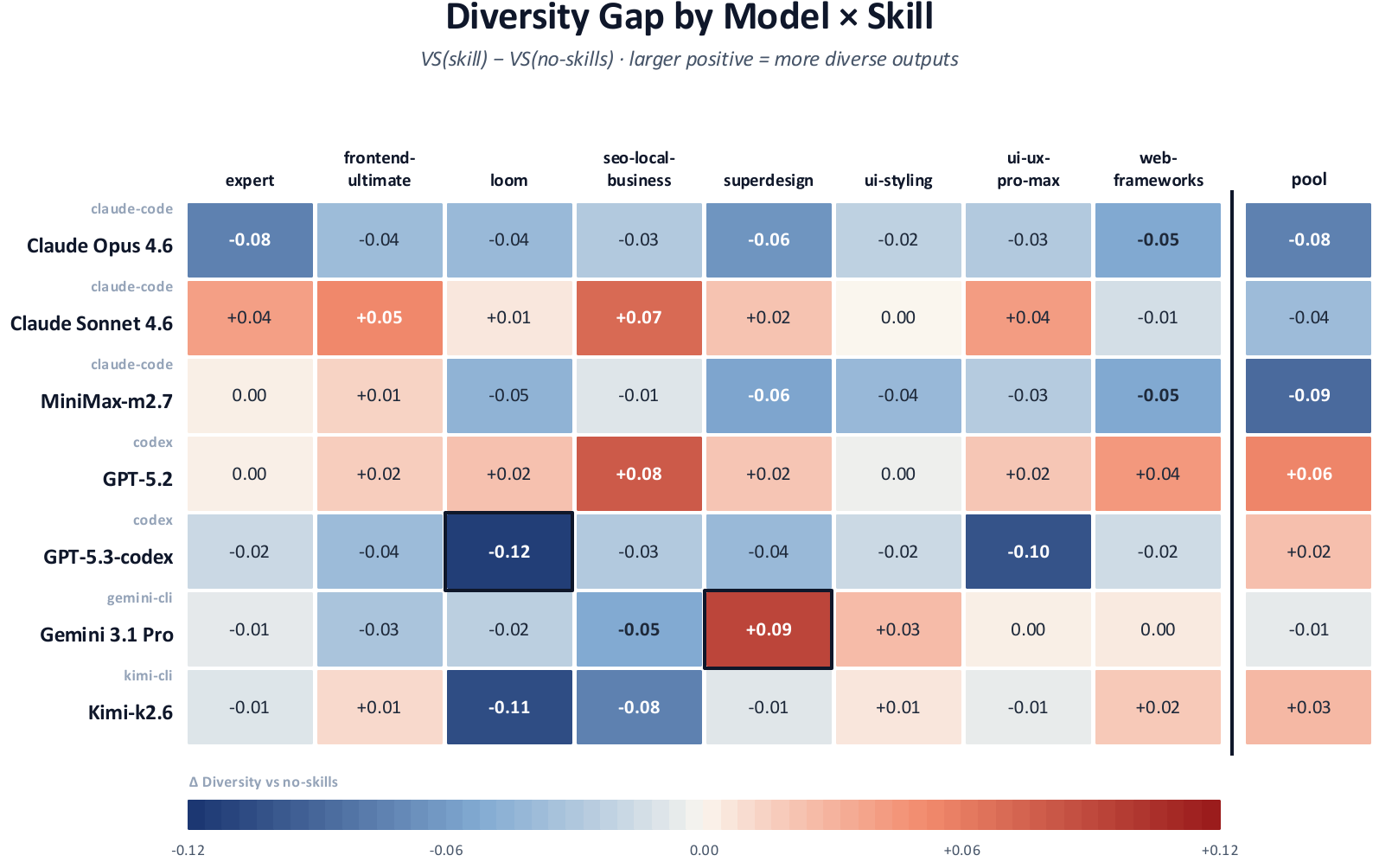}
\caption{Impact of web design skills on stylistic diversity relative to the \emph{no-skills} baseline, measured by changes in within-group Vendi Score computed from CSD-ViT-L style embeddings. Positive values indicate more diverse outputs under a given skill, while negative values indicate stronger stylistic convergence. The \emph{pool} column aggregates outputs across all skills for cross-skill analysis.}
\label{fig: impact3}
\end{figure*}

\FloatBarrier
\subsection{Evaluation Prompt} \label{appendix: evaluation prompt}

\begin{rubricbox}{1.\ Data Visualization}

\rubric{Insight Expression}{single image}
\begin{lstlisting}[style=promptText]
Evaluate the **insight expression** of this data visualization.

The visualization was created to convey a specific insight:

**Goal insight**: {insight}

Criteria:
- Does the chosen visualization type effectively communicate this insight?
- Can the reader **actually** understand the key message at a glance, as rendered?
- Are there effective annotations, highlights, or emphasis that draw attention to the insight?
- Is the presentation approach well-suited to the data characteristics?
- Does the visualization tell a clear story, or does the reader have to work to extract meaning?

Important: Judge the **end result as the reader would experience it**. A structurally sound approach that is undermined by severe rendering problems (e.g., overlapping labels hiding key annotations, illegible text in critical areas) should be scored lower -- the insight is only "expressed" if the reader can actually perceive it. Conversely, do NOT double-penalize minor cosmetic issues already covered by Visual Quality; only penalize rendering problems that **directly obstruct** the target insight.

Note: The creator was free to choose ANY chart type or visual approach. Do NOT penalize for unconventional choices -- judge ONLY whether the insight is effectively conveyed.

Scoring (1-5):
- 5: Reader grasps the key message immediately as rendered, clever use of annotation/emphasis/contrast, all insight components clearly visible
- 4: Insight clearly expressed, reasonable approach, minor gaps in emphasis or supporting elements
- 3: Insight basically understandable but presentation is generic or partially obscured, reader must interpret on their own
- 2: Presentation approach mismatches the insight goal, or key information is buried/illegible
- 1: Cannot read the target insight from the visualization

Return JSON: {"score": <1-5>, "reason": "<brief explanation>"}
\end{lstlisting}

\rubric{Data Accuracy}{single image + source text}
\begin{lstlisting}[style=promptText]
Evaluate the **data accuracy and factual fidelity** of this data visualization.

You are given the visualization, the original data, and background context. Check:
- Are the data points, proportions, and trends consistent with the source data?
- Are axis scales, labels, and values correct?
- Are any annotations or callouts traceable to the source material?
- Is there any fabricated data or misleading representation?

--- SOURCE DATA ---
{source_data}
--- END SOURCE DATA ---

--- SOURCE CONTEXT ---
{source_brief}
--- END SOURCE CONTEXT ---

Scoring (1-5):
- 5: All data points, proportions, and trends match source exactly; annotations traceable to source; no fabrication
- 4: Core data correct, minor deviations in secondary data points or slight extrapolation
- 3: Mostly correct, but notable numerical errors, proportion distortion, or untraceable annotations
- 2: Multiple data inconsistencies with source, or obvious fabricated content
- 1: Extensive data errors or fabrication, visualization is not trustworthy

Return JSON: {"score": <1-5>, "reason": "<brief explanation>"}
\end{lstlisting}

\rubric{Visual Quality}{single image}
\begin{lstlisting}[style=promptText]
Evaluate the **visual quality** of this data visualization.

Criteria:
- Color scheme: harmonious palette, appropriate for the topic, colorblind-friendly if applicable
- Layout: clean arrangement, proper spacing, nothing overlapping
- Labels and annotations: title, axis labels, legend, units -- all present and readable
- Typography: readable fonts, clear size hierarchy
- Professional finish: publication-ready quality, attention to detail

Scoring (1-5):
- 5: Harmonious colors, polished layout, complete labels (title/axes/legend/units), publication-grade professional quality
- 4: Good colors and layout, labels mostly complete, minor flaws (uneven spacing, label overlap)
- 3: Basic colors, functional layout but lacks design sophistication, some labels missing
- 2: Monotonous or clashing colors, rough layout, labels severely lacking
- 1: Chaotic styles, overlapping elements, visually unacceptable

Return JSON: {"score": <1-5>, "reason": "<brief explanation>"}
\end{lstlisting}

\rubric{Completeness}{single image}
\begin{lstlisting}[style=promptText]
Evaluate the **task completeness** of this data visualization.

The task requirements are provided below. Check whether the goal's insight is expressed and all requirements are met.

Task requirements:
{task_requirements}

Scoring (1-5):
- 5: Goal insight fully expressed, all requirements completely satisfied
- 4: Core requirements met, minor omissions (e.g., missing unit or a specific annotation)
- 3: Most requirements met, but notable gaps
- 2: Only partially satisfied
- 1: Key requirements not met, output unusable

Return JSON: {"score": <1-5>, "reason": "<brief explanation>"}
\end{lstlisting}

\end{rubricbox}

\begin{rubricbox}{2.\ Poster Generation }

\rubric{Design}{single image}
\begin{lstlisting}[style=promptText]
Evaluate the **visual design quality** of this poster/infographic.

Criteria:
- Color scheme: harmonious palette, appropriate for the topic and tone
- Layout: clean alignment, proper spacing, clear visual hierarchy
- Typography: readable fonts, clear size hierarchy (title > heading > body)
- Consistency: unified style throughout (colors, fonts, spacing)
- Professional polish: attention to detail, no overlapping elements

Scoring (1-5):
- 5: Harmonious colors, polished layout, clear hierarchy, professional design quality
- 4: Good colors and layout, minor flaws (slight misalignment, spacing issues)
- 3: Basic color scheme, layout functional but lacks design sophistication
- 2: Monotonous or clashing colors, rough layout
- 1: Chaotic styles, overlapping elements, hard to read

Return JSON: {"score": <1-5>, "reason": "<brief explanation>"}
\end{lstlisting}

\rubric{Completeness}{single image}
\begin{lstlisting}[style=promptText]
Evaluate the **task completeness** of this poster/infographic.

The task requirements are provided below. Check whether ALL requirements are met.

Task requirements:
{task_requirements}

Scoring (1-5):
- 5: All requirements fully satisfied
- 4: Core requirements met, minor omissions
- 3: Most requirements met, some notable gaps
- 2: Only partially satisfied
- 1: Key requirements not met

Return JSON: {"score": <1-5>, "reason": "<brief explanation>"}
\end{lstlisting}

\end{rubricbox}

\begin{rubricbox}{3.\ PPT Generation }

\rubric{Content}{per-slide image}
\begin{lstlisting}[style=promptText]
Evaluate the **content quality** of this presentation slide.

Judge how effectively this slide delivers its key message to the reader.

Criteria:
- Key message: does the slide have a clear takeaway that the reader can grasp?
- Information density: appropriate amount of content (not too crowded, not too sparse)
- Clarity: text is well-written, grammatically correct, easy to understand
- Text-visual balance: charts/images complement the text and reinforce the message

If rendering issues (truncated text, blank charts) cause information to be LOST, deduct proportionally to how much of the slide's message is affected. But if the core message still comes through despite minor visual flaws, do not over-penalize -- visual polish is scored under Design.

Scoring (1-5):
- 5: Clear key message, well-developed points, visuals and text complement each other
- 4: Key message clear, good content, but minor gaps (e.g., a chart lacks labels, text slightly sparse)
- 3: Core information present but message is diluted (e.g., missing title, no clear takeaway, or significant content lost to rendering)
- 2: Key message unclear, content poorly organized or mostly lost to rendering issues
- 1: No discernible message -- slide is blank, empty, or content entirely unreadable

Return JSON: {"score": <1-5>, "reason": "<brief explanation>"}
\end{lstlisting}

\rubric{Design}{per-slide image}
\begin{lstlisting}[style=promptText]
Evaluate the **visual design** of this presentation slide.

Criteria:
- Color scheme: harmonious, appropriate for the tone
- Layout: clean alignment, proper spacing, no overlapping elements
- Typography: readable fonts, clear hierarchy (title vs body)
- Visual elements: backgrounds, icons, shapes that enhance the message
- Consistency: matches the overall deck style

Scoring (1-5):
- 5: Harmonious colors, engaging visual elements, professional and polished
- 4: Good colors with some visual elements, minor design flaws
- 3: Basic color scheme, with rough layout and no supplementary visual elements 
- 2: Monotonous black/white, readable but unappealing
- 1: Conflicting styles, content difficult to read

Return JSON: {"score": <1-5>, "reason": "<brief explanation>"}
\end{lstlisting}

\rubric{Completeness}{full deck images}
\begin{lstlisting}[style=promptText]
Evaluate the **task completeness** of this presentation.

Check whether the required CONTENT is present in the slides. For each requirement below, determine if the corresponding content exists in the presentation.

Task requirements:
{task_requirements}

Important:
- Judge whether the required content IS PRESENT, not how it looks visually.
- A chart that renders blank or empty counts as MISSING content (deduct).
- Text that is overlapping or hard to read but IS present counts as COMPLETE (do not deduct -- visual issues are scored under Design).

Scoring (1-5):
- 5: All required content present -- every section, data point, and chart accounted for
- 4: Core content present, minor omissions (e.g., one data point or detail missing)
- 3: Most content present, but some required sections or charts are missing
- 2: Only partially complete -- multiple required elements missing
- 1: Key required content not present

Return JSON: {"score": <1-5>, "reason": "<brief explanation>"}
\end{lstlisting}

\end{rubricbox}

\begin{rubricbox}{4.\ Report Generation}

\rubric{Content Quality}{report text only}
\begin{lstlisting}[style=promptText]
Evaluate the **content quality** of this report across two aspects: writing quality AND analysis depth.

A. Writing & Structure:
- Organization: clear headings, logical flow, well-structured executive summary
- Clarity: well-written, grammatically correct, easy to understand
- Information density: appropriate amount of content (not too sparse, not overwhelming)
- Logical consistency: no self-contradicting claims or misleading interpretations

B. Analysis Depth:
- Are insights substantive (backed by statistics specific data) or surface-level (just restating numbers)?
- Does the analysis go beyond descriptive statistics (e.g., correlation, attribution, comparison)?
- Are conclusions and recommendations data-driven and specific (not generic advice)?
- Is there unnecessary redundancy (e.g., same data presented in multiple formats without adding value)?

Scoring (1-5):
- 5: Excellent writing AND deep analysis -- insights are substantive, data-driven, logically consistent, with no redundancy
- 4: Good writing and organization, but some sections lack analytical depth or have minor redundancy
- 3: Adequate structure but analysis is mostly surface-level, or notable redundancy/inconsistency
- 2: Poor organization, stiff writing, and shallow analysis
- 1: Chaotic structure, grammar errors, no real analysis

Return JSON: {"score": <1-5>, "reason": "<brief explanation>"}
\end{lstlisting}

\rubric{Visualization}{full-page overview + scrolled crops}
\begin{lstlisting}[style=promptText]
Evaluate the **visualization quality** of this report.

Criteria:
- Chart type selection: appropriate for the data (bar for comparison, line for trend, pie for composition, etc.)
- Color scheme: professional, harmonious, accessible
- Labels and annotations: axis titles, data labels, legends, units
- Readability: data is immediately clear from the visualization
- Tables: properly formatted, aligned, with clear headers

Scoring (1-5):
- 5: Chart types perfectly matched to data, professional colors, complete annotations, data immediately clear
- 4: Good chart types, has titles and labels, good visual effect
- 3: Basic charts present, but type selection or labels could be improved
- 2: Few charts or wrong types, missing labels
- 1: No charts or charts are incomprehensible

Return JSON: {"score": <1-5>, "reason": "<brief explanation>"}
\end{lstlisting}

\rubric{Completeness}{report text only}
\begin{lstlisting}[style=promptText]
Evaluate the **task completeness** of this report.

Go through EVERY requirement below and check whether it is substantively addressed in the report. Do NOT just check if a section heading exists -- verify that the actual content within each section fulfills what was asked. For example, if a KPI is required "segmented by genre and year", check that BOTH segmentations are present, not just one.

Task requirements:
{task_requirements}

Scoring (1-5):
- 5: Every requirement substantively addressed -- all sections have the requested content, all KPIs present with required segmentation
- 4: Core requirements met, but 1-2 minor details missing (e.g., a KPI lacks one segmentation dimension)
- 3: Most requirements met, but notable gaps (e.g., a section is present but missing key requested analysis)
- 2: Only partially satisfied -- multiple sections lack required content
- 1: Key requirements not met

Return JSON: {"score": <1-5>, "reason": "<brief explanation>"}
\end{lstlisting}

\end{rubricbox}

\begin{rubricbox}{5.\ Web Design }

\rubric{Visual Design}{per-page multi-image (full + crops)}
\begin{lstlisting}[style=promptText]
Evaluate the **visual design execution quality** of this web page.

Criteria:
- Color & typography: harmonious palette, readable fonts, clear heading hierarchy (h1 > h2 > body), consistent font sizing
- Layout & structure: well-organized sections, clear information hierarchy, consistent grid alignment, no misaligned or jagged elements
- Spacing & polish: appropriate white space between sections, no elements overlapping or clipped, no exposed HTML tags or rendering artifacts
- Placeholder handling: the task does NOT provide image/video assets, so placeholder elements are expected. Do NOT penalize for the absence of real images. Instead, evaluate whether placeholders are well-designed -- appropriate size and position, consistent styling with the overall theme (e.g., labeled gray boxes are better than empty black voids)

Note: the first image is a fullpage overview, subsequent images are detail crops scrolled top-to-bottom. Boundary cuts between detail images are screenshot artifacts.

Scoring (1-5):
- 5: Harmonious color palette with intentional accent choices; typography has clear h1/h2/body hierarchy with consistent sizing; sections are well-separated with balanced spacing; grid alignment is pixel-consistent; placeholders are styled to match the theme (e.g., colored boxes with labels, icon placeholders)
- 4: Color palette is coordinated, typography hierarchy is clear, layout is well-organized; has 1-2 minor issues such as: one section has slightly uneven spacing, a placeholder is slightly oversized, a font weight is inconsistent in one area
- 3: Has a color scheme and basic section structure, but execution is noticeably rough: spacing is uneven across sections, heading sizes are inconsistent, placeholders are unstyled empty blocks, or the page feels like a wireframe with colors applied
- 2: Multiple visible problems: elements overlap or get clipped, sections are misaligned, exposed HTML tags or broken CSS visible, color choices clash, or large portions of the page are visually broken
- 1: Page renders as unstyled HTML (no CSS applied), or the page fails to load / displays a blank screen

Return JSON: {"score": <1-5>, "reason": "<brief explanation>"}
\end{lstlisting}

\rubric{Responsive}{per-device screenshots (mobile / tablet)}
\begin{lstlisting}[style=promptText]
Evaluate the **responsiveness** of this website on a {device_name}.

You are shown screenshots captured with {device_description}.
Each screenshot is from a different page of the same website. Evaluate the overall responsive quality across all pages shown.

Criteria:
- No horizontal scrollbar / content overflow
- Navigation is accessible (hamburger menu or adapted nav)
- Touch targets are appropriately sized (>=44px)
- Text is readable without zooming
- Content adapts to the viewport width (no desktop layout forced into a smaller screen)

Note: the task does NOT provide image/video assets. Pages may contain elements explicitly labeled or styled as placeholders (e.g., boxes with "placeholder" text, icons indicating missing media). Do NOT penalize these as a responsive issue -- only evaluate how the layout, navigation, and content elements adapt to this viewport.

Scoring (1-5):
- 5: No overflow, navigation properly adapted (e.g., hamburger menu on mobile), touch targets well-sized, content reflows cleanly to viewport width
- 4: Good adaptation overall; 1-2 minor issues such as: nav items slightly tight, one button partially clipped at edge, or slight spacing inconsistency
- 3: Has viewport meta and attempts responsive layout, but noticeable problems: nav not collapsed, some text requires horizontal scroll, or several elements are cramped
- 2: Layout significantly broken: content overflows viewport, navigation is unusable, or desktop layout is forced into the smaller screen
- 1: No responsive handling at all -- page renders at desktop width requiring zoom and scroll

Return JSON: {"score": <1-5>, "reason": "<brief explanation>"}
\end{lstlisting}

\end{rubricbox}


\end{document}